\documentclass[pdflatex,sn-mathphys-num,iicol]{sn-jnl}

\usepackage{microtype}
\usepackage{graphicx}
\usepackage[table,xcdraw]{xcolor}
\usepackage{multirow}
\usepackage{colortbl}
\usepackage{booktabs}
\usepackage{makecell}
\usepackage{tabularx}
\usepackage{array}
\usepackage{bbding}
\usepackage{bm}
\usepackage{enumitem}
\usepackage{placeins}
\usepackage{cuted}
\usepackage{amsmath,amssymb,amsfonts,mathtools}
\usepackage{textcomp}
\usepackage{pifont}
\usepackage{float}
\usepackage[capitalize,noabbrev]{cleveref}
\usepackage{xurl}

\makeatletter
\def\email#1{\global\advance\emailcnt by 1\relax%
\if@corauemail%
  \g@addto@macro\corrauthemail{%
  \setcounter{footnote}{0}%
  \textcolor{blue}{#1}.\ %
  }%
\else%
  \g@addto@macro\authemail{%
  \setcounter{footnote}{0}%
  \textcolor{blue}{#1}.\ %
  }%
\fi}
\makeatother

\graphicspath{{fig/}}

\definecolor{brickred}{rgb}{0.8, 0.0, 0.0}
\definecolor{oceanblue}{RGB}{0, 0, 128}
\definecolor{best}{RGB}{232, 241, 238}
\definecolor{second}{rgb}{0.85, 0.9, 1.0}
\definecolor{rowred}{rgb}{1,0.9,0.9}
\definecolor{purple}{rgb}{230,230,255}
\definecolor{tableblue}{HTML}{BEE2FF}
\definecolor{naturedraft}{RGB}{132,39,57}
\definecolor{refblue}{RGB}{0,63,125}
\definecolor{citeblue}{RGB}{0,83,120}

\hypersetup{
  hypertexnames=false,
  colorlinks=true,
  citecolor=citeblue,
  linkcolor=refblue,
  urlcolor=refblue
}
\makeatletter
\let\table\tableorg
\let\endtable\endtableorg
\makeatother

\theoremstyle{thmstyleone}

\theoremstyle{thmstylethree}

\theoremstyle{thmstyletwo}

\begin{document}

\title[FreeCam]{Rethinking Camouflage Image Generation towards a Training-Free Paradigm}

\author[1]{\fnm{Haodong} \sur{Yang}}
\equalcont{These authors contributed equally to this work.}
\author[1]{\fnm{Zhongling} \sur{Huang}}
\equalcont{These authors contributed equally to this work.}
\author*[1]{\fnm{Gong} \sur{Cheng}}\email{chenggong1119@gmail.com}

\affil[1]{\orgdiv{School of Automation}, \orgname{Northwestern Polytechnical University}, \orgaddress{\city{Xi'an}, \country{China}}}

\abstract{
% Camouflage Image Heneration (CIG) aims to synthesize realistic images in which a preserved foreground target becomes less perceptually separable from semantically compatible surroundings.
% Camouflage image generation aims to synthesize realistic camouflaged images by preserving foreground objects while generating semantically compatible surroundings and promoting appearance assimilation between the foreground and background.
% Camouflage Image Generation (CIG) aims to synthesize realistic camouflaged images by generating semantically compatible and visually coherent surroundings around the foreground. 
% Camouflage image generation aims to synthesize realistic camouflaged images by blending foreground objects into diverse background contexts. Achieving this objective requires jointly satisfying three coupled requirements: preserving the foreground target, placing it in a context that is semantically plausible and favorable to concealment, and reducing its appearance discrepancy from the surrounding background. Existing approaches address these requirements through task-specific optimization on camouflaged object datasets, either by directly adapting foreground appearance to a given background or by fine-tuning diffusion models to synthesize concealing surroundings conditioned on the foreground.

Camouflage image generation (CIG) aims to synthesize realistic camouflaged images by blending foreground objects into concealment-compatible background contexts. Achieving this objective requires jointly satisfying three coupled requirements: foreground preservation to retain target integrity, semantic compatibility to select plausible concealment contexts, and appearance assimilation to reduce visual discrepancies.
Recent approaches predominantly rely on task-specific training on camouflage datasets to address these requirements, incurring substantial computational cost and limiting generalization beyond the training domain.
To address these limitations, we formulate \emph{training-free CIG} as a concealment-oriented paradigm that preserves the target while reducing its perceptual separability from the synthesized surroundings, rather than maintaining its visual prominence, without parameter updates.
We instantiate this paradigm with FreeCam based on a frozen inpainting diffusion framework to preserve the foreground. Within this framework, a Contextual Reasoning Module exploits frozen multimodal priors to infer an environment favorable to concealment, thereby promoting semantic compatibility, while an Intrinsic Appearance Module extracts low-level color and texture cues from the foreground to guide background synthesis toward appearance assimilation.
Extensive experiments demonstrate that FreeCam achieves state-of-the-art generation quality and camouflage effectiveness without task-specific training, while its generated images provide synthetic supervision for camouflaged object detection and reduce target detectability under general object detectors.

}


\maketitle

\begin{figure*}[!t]
\centering
\includegraphics[width=0.98\linewidth]{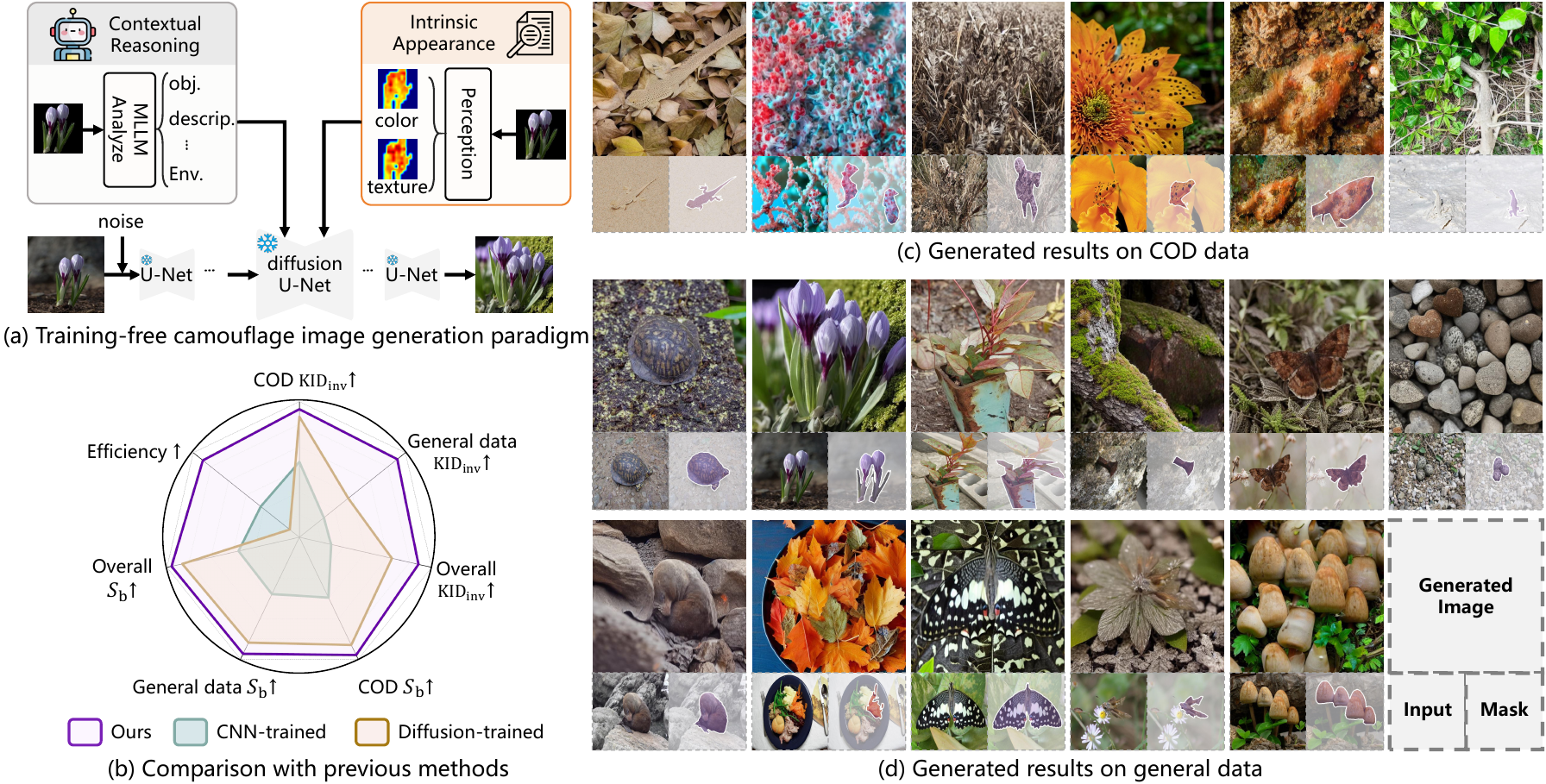}
\caption{(a) Illustration of the proposed training-free CIG paradigm, in which contextual reasoning and intrinsic appearance conditioning guide a frozen diffusion model toward semantic compatibility and appearance assimilation. (b) Comparison with CNN-trained and diffusion-trained CIG methods across seven normalized axes: efficiency; inverted KID on COD, general data, and the overall set; and $S_{\mathrm{b}}$ on the same three evaluation sets. All axes are oriented such that larger values are better. (c--d) The generated results on (c) COD and (d) general data validate the capability of the proposed paradigm.}
\label{fig:1}
\end{figure*}

\begin{figure*}[!t]
\centering
\includegraphics[width=0.95\linewidth]{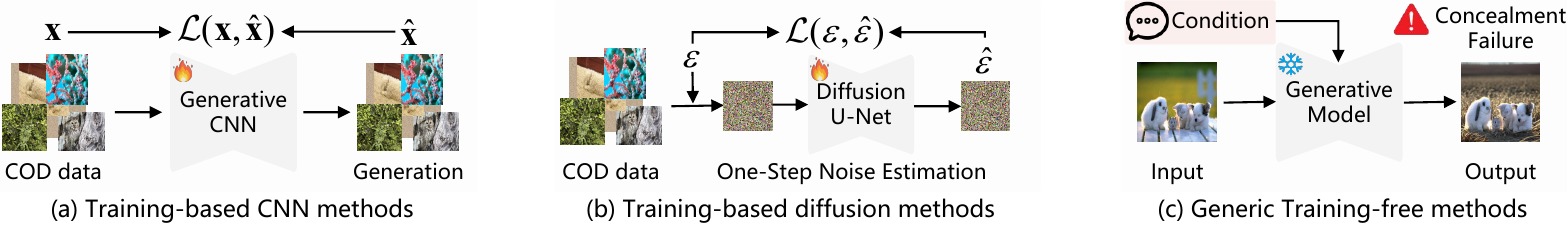}
\caption{(a--b) Training-based CNN and diffusion methods rely on COD-specific optimization. (c) Generic training-free generation steers frozen priors and avoids task-specific optimization costs but is not concealment-oriented.}
\label{fig:moti}
\end{figure*}

\section{Introduction}
\label{sec:intro}

% Camouflaged objects pose a fundamental challenge to visual perception by blending into their surrounding contexts~\cite{zhao2025deep}, and accurately modeling such concealment is crucial for safety-critical applications, including pest control~\cite{ebrahimi2017vision} and healthcare~\cite{ji2024rethinking, yuan2023full}.
% However, the construction of Camouflage Object Datasets (CODs) remains severely constrained: the inherently low saliency of camouflaged targets makes them difficult to discover and capture, resulting in limited category diversity~\cite{le2019anabranch} and labor-intensive pixel-level annotations~\cite{fan2020camouflaged}.
% To alleviate this bottleneck, Camouflage Image Generation (CIG)~\cite{CI} has emerged as a scalable alternative for synthesizing high-quality camouflage imagery~\cite{zhao2024lake,das2025camouflage}.
% Broadly, CIG aims to synthesize realistic camouflaged images by blending foreground objects into diverse background contexts.
% Despite recent progress, CIG still faces a fundamental tension between \emph{efficiency} and \emph{generalization} because domain-specific training is computationally costly, undermining efficiency, while such task-specific optimization induces overfitting to particular camouflage domains and limits generalization to general data, as illustrated in Fig.~\ref{fig:1}(b).

Camouflaged objects pose a persistent challenge to visual perception, since the visual cues that typically distinguish an object from the background, such as color contrast and texture discrepancy, are suppressed or absorbed by the surrounding contexts~\cite{zhao2025deep}. Faithfully modeling such concealment is important for visual understanding tasks where targets are intrinsically inconspicuous, including agricultural pest monitoring~\cite{ebrahimi2017vision,wu2025knowledge,zhang2022agripest,song2025rdw} and healthcare-related visual analysis~\cite{ji2024rethinking,yuan2023full,xiao2023concealed,fan2021concealed}. However, constructing large-scale Camouflaged Object Datasets (CODs) remains inherently difficult: camouflaged targets are hard to discover and capture in natural scenes due to their low saliency, resulting in limited category and scene diversity. Even after collection, reliable pixel-level annotation still requires substantial manual effort. To alleviate this bottleneck, Camouflage Image Generation (CIG) has become a scalable route for synthesizing camouflaged imagery by assimilating foreground objects into semantically compatible and visually coherent background contexts~\cite{zhao2024lake,das2025camouflage,CTCIG}. Unlike conventional image generation, which typically seeks to make the foreground recognizable and faithfully rendered, CIG instead treats perceptual concealment as the generation objective, making it a distinct and more constrained problem.

This pursuit of concealment imposes three coupled requirements on CIG: \emph{foreground preservation}, \emph{semantic compatibility}, and \emph{appearance assimilation}. Foreground preservation requires retaining the target's semantic identity and visual integrity during synthesis. Semantic compatibility requires the synthesized surroundings to be both plausible for the target and favorable to effective concealment. Appearance assimilation requires suppressing perceptual discrepancies between the target and its local surroundings, particularly in color distribution and texture patterns. An effective CIG method is therefore required to preserve the target in the final image while reducing its perceptual separability from semantically compatible and visually coherent surroundings, whereas conventional image generation approaches typically maintain or enhance target distinctiveness.

% This pursuit of concealment imposes three coupled requirements on CIG: \emph{foreground preservation}, \emph{semantic compatibility}, and \emph{appearance assimilation}. Foreground preservation requires retaining the target's semantic identity and visual integrity. Semantic compatibility requires the synthesized surroundings to be both plausible for the target and favorable to effective concealment. Appearance assimilation requires suppressing perceptual discrepancies between the target and its local surroundings, particularly in color distribution and texture patterns. An effective CIG method is therefore required to preserve the target in the final image while reducing its perceptual separability from semantically compatible and visually coherent surroundings, whereas conventional image generation typically maintains or enhances target distinctiveness.

Existing CIG methods can be broadly grouped into convolutional neural network-based (CNN-based) and diffusion model-based (DM-based) approaches. As illustrated in Fig.~\ref{fig:moti}(a), CNN-based methods~\cite{DCI,LCGNet} pursue appearance assimilation by learning a supervised image-to-image camouflage mapping from COD, but may compromise foreground preservation. More recently, DM-based methods~\cite{zhao2024lake,chen2025foreground,das2025camouflage,CTCIG} formulate CIG as conditional generation and fine-tune diffusion models on COD through task-specific denoising objectives to implicitly address the above three requirements, as illustrated in Fig.~\ref{fig:moti}(b). Despite their technical differences, both paradigms remain dependent on COD supervision to capture camouflage-specific generation patterns. This training paradigm not only incurs expensive optimization and limits deployment efficiency, but also couples the generator to COD-specific representations, thereby restricting its generalization and applicability to general-domain scenarios, as illustrated in Fig.~\ref{fig:1}(b). These limitations raise a more fundamental question: can pretrained generative priors be steered at inference time to synthesize realistic camouflaged images without COD-specific optimization while retaining their broad generative capability?

Recent advances in training-free image generation offer a possible route to this question: pretrained diffusion models encode broad generative priors that can be controlled at inference time through external conditions~\cite{xia2025tf,mao2025tuning}, as illustrated in Fig.~\ref{fig:moti}(c). Existing approaches mainly adopt prompt-driven~\cite{chen2025t2i}, reference-conditioned~\cite{zheng2025lanpaint}, and hybrid~\cite{mao2025tuning,qin2025free} designs, which steer the frozen prior through text prompts, reference images, or both. However, these mechanisms are designed for conventional generation, where control typically preserves subject identity while maintaining or enhancing visual prominence, rather than reducing the subject's perceptual separability from its surroundings. Consequently, \emph{semantic plausibility} alone does not ensure compatibility with concealment: placing a fish in water is semantically reasonable, yet the generated scene may still leave the target visually conspicuous. 
Moreover, even when reference images are provided, these controls typically prioritize \emph{appearance fidelity}, preserving target-specific color and texture characteristics and thereby maintaining visual distinctiveness.
Taken together, these limitations indicate that generic training-free controls are not designed to jointly satisfy the three requirements of CIG, motivating a dedicated training-free formulation for concealment-oriented synthesis.

% Moreover, these controls do not explicitly regulate the local color and texture relations required for camouflage, even when reference images are provided, leaving \emph{appearance assimilation} insufficiently addressed. Taken together, these limitations indicate that generic training-free controls are not designed to jointly satisfy the three requirements of CIG, motivating a dedicated training-free formulation for concealment-oriented synthesis.

In this paper, we formulate \emph{training-free CIG} as a paradigm for reducing the perceptual separability of a preserved foreground target from its synthesized surroundings through inference-time steering of frozen generative priors. Unlike generic training-free generation, realizing this paradigm requires addressing two camouflage-specific challenges: \emph{semantic compatibility} for identifying \emph{which} contexts are plausible for the target and favorable to concealment, and \emph{appearance assimilation} for determining \emph{how} to reduce visual discrepancies between the synthesized surroundings and the target. We instantiate this paradigm with FreeCam, as illustrated in Fig.~\ref{fig:1}(a), which addresses these two challenges through a Contextual Reasoning Module (CRM) and an Intrinsic Appearance Module (IAM), respectively. The CRM leverages multimodal reasoning priors to infer an environment favorable to concealment and encodes it as a semantic condition, thereby promoting \emph{semantic compatibility}. The IAM extracts low-level characteristics of the foreground, particularly color and texture cues, and converts them into an appearance condition that guides background synthesis toward \emph{appearance assimilation}. These conditions jointly steer a frozen inpainting diffusion model to generate only the background region, thereby preserving the original foreground by construction.

We evaluate FreeCam on the LAKE-RED benchmark~\cite{zhao2024lake} against representative training-based CNN and diffusion methods, and use these comparisons to assess the feasibility of training-free CIG. Across the camouflaged, salient, and general subsets, FreeCam variants achieve state-of-the-art (SOTA) generation quality in terms of KID without COD-specific optimization. 
Results on the boundary score $S_{\mathrm{b}}$ further indicate that FreeCam produces less distinguishable object boundaries, providing additional evidence of improved camouflage effectiveness.
Fig.~\ref{fig:1}(b) summarizes the overall comparison in terms of generation quality, camouflage effectiveness, and efficiency. Downstream augmentation experiments further show that FreeCam-generated samples improve the performance of camouflaged object detectors, demonstrating their utility as synthetic training data. Visual results on inputs from both camouflage and general domains, shown in Fig.~\ref{fig:1}(c--d), further illustrate the semantic coherence and appearance assimilation achieved by FreeCam.

The main contributions of this work are summarized as follows:
\begin{itemize}
    \item We formulate \emph{training-free camouflage image generation} as a concealment-oriented paradigm whose objective differs from that of generic training-free generation: rather than maintaining or enhancing the target's visual prominence, it seeks to reduce the perceptual separability between a preserved target and its synthesized surroundings.

    \item We identify two camouflage-specific design principles for realizing this paradigm: \emph{semantic compatibility} for selecting \emph{which} contexts are plausible for the target and favorable to concealment, and \emph{appearance assimilation} for determining \emph{how} to reduce visual discrepancies between the target and the synthesized surroundings.

    \item We propose \textbf{FreeCam} as an instantiation of this paradigm. FreeCam preserves the original foreground by restricting synthesis to the background region, while a Contextual Reasoning Module addresses \emph{semantic compatibility} by inferring an environment favorable to concealment and encoding it as a semantic condition, and an Intrinsic Appearance Module addresses \emph{appearance assimilation} by extracting foreground color and texture cues as an appearance condition.

    \item Extensive experiments evaluate generation quality, camouflage effectiveness, computational efficiency, and downstream perception across diverse evaluation settings. FreeCam achieves consistent gains in generalization and efficiency over state-of-the-art methods while serving as an efficient data engine for camouflage image generation, demonstrating the feasibility and scope of training-free CIG.
\end{itemize}

\section{Related Work}

\subsection{Camouflage Image Generation}
CIG aims to synthesize images in which a foreground target is concealed by its surroundings while preserving its semantic identity and visual integrity. Existing approaches can be broadly divided into CNN-based and DM-based methods. Early work~\cite{CI} explored this task using handcrafted features to composite a foreground object onto a given background and reduce its perceptual saliency to human observers. Building on this formulation, CNN-based methods~\cite{DCI,LCGNet} employ deep networks to adapt the foreground appearance to a given background through style transfer and feature fusion. These methods primarily promote appearance assimilation, but their direct modification of the target may compromise foreground preservation. More recently, DM-based methods~\cite{zhao2024lake,chen2025foreground,das2025camouflage,CTCIG} have become the dominant paradigm. Rather than requiring a predefined background, these methods preserve the input foreground and synthesize its surrounding context. Leveraging the generative capacity of DMs, they formulate CIG as supervised conditional generation and fine-tune pretrained DMs on CODs to implicitly learn semantic compatibility and appearance assimilation.

Despite their effectiveness, learning-based methods across both paradigms generally require COD-specific optimization. This training paradigm incurs substantial computational cost and biases the learned generator toward camouflage-specific data distributions, which can restrict generalization beyond the training domain. 
As a result, whether CIG can preserve the foreground while establishing semantic compatibility and appearance assimilation without COD-specific training remains largely unexplored. This work addresses this gap by formulating training-free CIG as a paradigm and instantiating it with \textbf{FreeCam}.
% As a result, whether the three CIG requirements can be jointly addressed without COD-specific training remains largely unexplored. In contrast, this work explores CIG from a training-free perspective and proposes \textbf{FreeCam}, which addresses this gap by steering frozen generative priors at inference time without task-specific parameter updates.

\subsection{Training-free Image Generation}
Recent advances in large-scale pretrained diffusion models~\cite{LDM,sdxl} have enabled a growing class of training-free generation methods that reuse frozen generative priors and introduce control at inference time without task-specific model training. These methods can be broadly categorized into prompt-driven, reference-conditioned, and hybrid designs according to the conditioning signals. Prompt-driven methods~\cite{chen2025t2i,zhang2024vascar,ma2025training} use semantic descriptors to specify scene content and composition, primarily providing semantic plausibility. Reference-conditioned methods~\cite{zheng2025lanpaint} impose visual constraints from reference images to preserve subject identity or appearance fidelity. Hybrid methods~\cite{li2024tuning,ding2024freecustom,avrahami2025stable} combine textual and visual signals to coordinate semantic and appearance control.
Collectively, these studies demonstrate that frozen diffusion priors can be repurposed through distinct conditions for diverse controllable synthesis tasks without task-specific parameter updates.
\begin{figure*}[!t]
\centering
\includegraphics[width=0.98\linewidth]{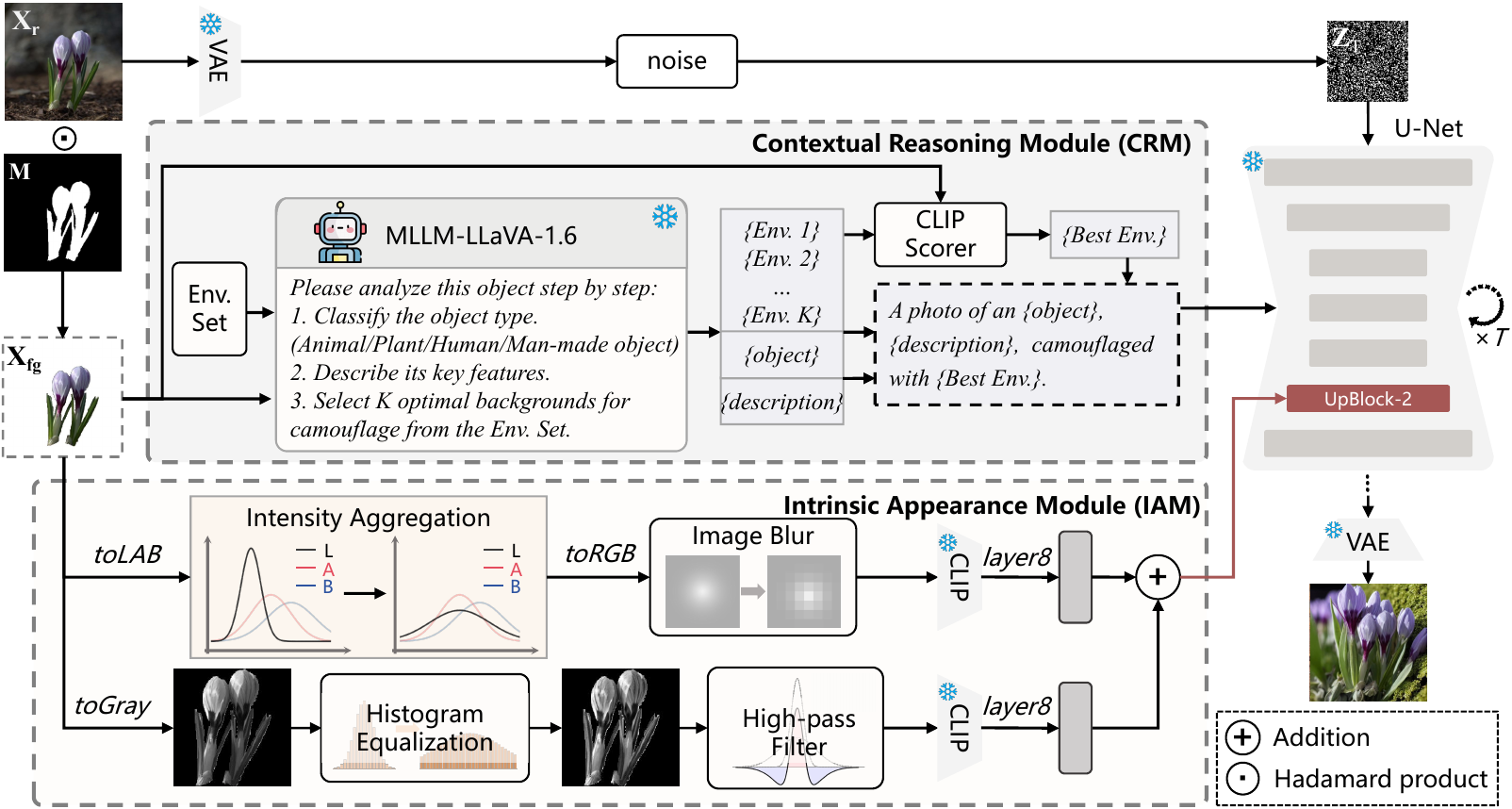}
\caption{Framework of FreeCam. Background-only inpainting preserves the original foreground. CRM addresses semantic compatibility by constructing a concealment-oriented semantic condition, whereas IAM addresses appearance assimilation by extracting foreground color and texture cues as an appearance condition. The two conditions jointly guide the frozen diffusion model during background synthesis.
% Framework of FreeCam.
% A frozen diffusion model is guided by contextual reasoning and low-level intrinsic appearance cues.
% The contextual reasoning module determines semantically suitable camouflage contexts from the foreground object, while the intrinsic appearance module extracts appearance cues that improve foreground--background consistency.
% Their integration steers generation toward semantically appropriate and perceptually coherent camouflage images without task-specific training.
}
\label{fig:method}
\end{figure*}

Despite their effectiveness in controllable synthesis, these mechanisms are not aligned with the concealment objective of CIG. Prompt-driven control can establish semantic plausibility, but a plausible scene is not necessarily favorable to concealment. Reference-conditioned control prioritizes appearance fidelity, which can preserve target-specific characteristics while maintaining visual distinctiveness from the surroundings. Hybrid methods inherit this objective mismatch despite combining both conditions. Consequently, generic training-free generation does not address the semantic compatibility and appearance assimilation required by training-free CIG. FreeCam instead derives semantic prompts specifying environments favorable to concealment and extracts intrinsic foreground appearance cues to promote assimilation, thereby aligning inference-time control with the concealment objective.

\section{Method}

\subsection{Problem Formulation}
\label{sec:formulation}

We instantiate the training-free CIG paradigm with \textbf{FreeCam}, as illustrated in Fig.~\ref{fig:method}. By restricting synthesis to the background region, FreeCam preserves the original foreground by construction. The remaining challenges are to establish semantic compatibility in context selection and appearance assimilation in color and texture. FreeCam addresses them through a \textbf{Contextual Reasoning Module (CRM)}, which constructs a semantic condition for selecting contexts favorable to concealment, and an \textbf{Intrinsic Appearance Module (IAM)}, which derives an appearance condition from foreground color and texture cues.

% We instantiate the training-free CIG paradigm with \textbf{FreeCam}, as illustrated in Fig.~\ref{fig:method}. FreeCam addresses the three coupled requirements of CIG through a background inpainting strategy and two guidance branches derived from the foreground, without any parameter updates. By restricting synthesis to the background region, FreeCam preserves the original foreground. Meanwhile, a \textbf{Contextual Reasoning Module (CRM)} derives a semantic condition that identifies contexts favorable to concealment, while an \textbf{Intrinsic Appearance Module (IAM)} extracts color and texture cues to promote appearance assimilation.

% FreeCam steers a frozen inpainting diffusion model with a semantic-appearance dual conditioning: (1) a \textbf{Contextual Reasoning Module (CRM)} that derives semantically compatible textual prompts, and (2) an \textbf{Intrinsic Appearance Module (IAM)} that extracts object intrinsic appearance cues to promote foreground-background assimilation.

Given a reference image
$\mathbf X_{\mathrm r}\in\mathbb R^{H\times W\times 3}$
and a binary foreground mask
$\mathbf M_{\mathrm{fg}}\in\{0,1\}^{H\times W}$,
we extract the foreground as
$\mathbf X_{\mathrm{fg}}
=
\mathbf X_{\mathrm r}\odot\mathbf M_{\mathrm{fg}}$,
where $\odot$ denotes the element-wise (Hadamard) product.
The corresponding inpainting mask is defined as
$\mathbf M_{\mathrm{inp}}
=
1-\mathbf M_{\mathrm{fg}}$,
such that synthesis is restricted to the background region.
The objective is to generate a composite image
$\mathbf I\in\mathbb R^{H\times W\times 3}$
that preserves the original foreground throughout the generation process while reducing its perceptual separability from the generated background through semantic compatibility and appearance assimilation.

The background generation is performed in the latent space of the pretrained variational autoencoder (VAE) in the frozen inpainting diffusion model~\cite{LDM}. 
CRM and IAM process the foreground in parallel to produce a semantic condition
$\mathbf C_{\mathrm{sem}}$
and an appearance condition
$\mathbf C_{\mathrm{app}}$,
respectively.
At denoising timestep $t$, the frozen diffusion model predicts the noise conditioned on the standard inpainting inputs and the two guidance conditions:
\begin{equation}
\widehat{\boldsymbol{\epsilon}}_t
=
\epsilon_\theta
\left(
\mathbf z_t,
t,
\mathbf C_{\mathrm{inp}},
\mathbf C_{\mathrm{sem}},
\mathbf C_{\mathrm{app}}
\right),
\label{eq:noise_est}
\end{equation}
where $\mathbf C_{\mathrm{inp}}$ denotes the standard inputs of the inpainting model, including $\mathbf M_{\mathrm{inp}}$ resized to the latent resolution and the latent encoding of $\mathbf X_{\mathrm{fg}}$, while $\mathbf z_t$ denotes the noisy latent at timestep $t$.

The semantic and appearance conditions are jointly incorporated into the frozen cross-attention layers, where their projected key and value tokens are concatenated to form the fused conditioning representation:
\begin{equation}
\begin{split}
\mathbf K_{\mathrm{fused}}
&=
\operatorname{concat}\!\left(
\mathbf W_K(\mathbf C_{\mathrm{sem}}),
\mathbf W_K(\mathbf C_{\mathrm{app}})
\right),\\
\mathbf V_{\mathrm{fused}}
&=
\operatorname{concat}\!\left(
\mathbf W_V(\mathbf C_{\mathrm{sem}}),
\mathbf W_V(\mathbf C_{\mathrm{app}})
\right),
\end{split}
\label{eq:fusion}
\end{equation}
where $\mathbf W_K$ and $\mathbf W_V$ denote the frozen key and value projection layers of the corresponding cross-attention block.
After denoising, the final latent is decoded by the VAE to obtain $\widehat{\mathbf X}$.
The final image is then constructed as
\begin{equation}
\mathbf I
=
\mathbf X_{\mathrm r}\odot\mathbf M_{\mathrm{fg}}
+
\widehat{\mathbf X}\odot\mathbf M_{\mathrm{inp}},
\label{eq:composition}
\end{equation}
thereby preserving the original foreground and using the decoded content only in the background region.
Detailed descriptions of CRM and IAM are provided in Secs.~\ref{sec:crm} and~\ref{sec:ism}, respectively.

\begin{figure}[!t]
\centering
\includegraphics[width=0.98\linewidth]{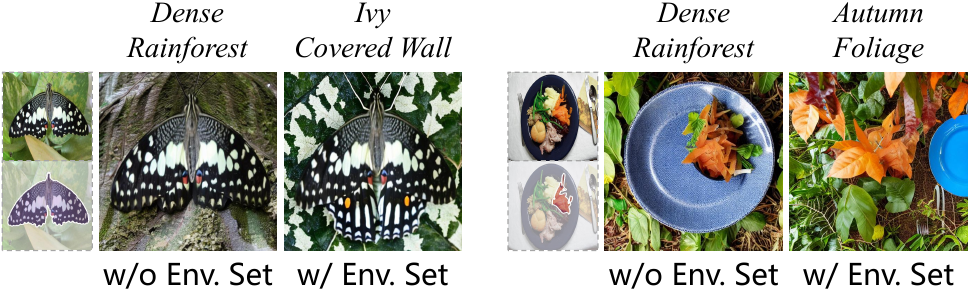}
\caption{Visual comparison of images generated with and without the environment set. The text above each image denotes the inferred environment.}
\label{fig:env}
\end{figure}

\subsection{Contextual Reasoning Module}
\label{sec:crm}

Existing training-based CIG methods~\cite{zhao2024lake,chen2025foreground,das2025camouflage} learn associations between foreground objects and camouflage contexts through optimization on CODs. In the training-free paradigm, however, such task-specific associations are unavailable. A general multimodal large language model (MLLM) provides broad semantic knowledge, but it may favor environments that are semantically plausible or ecologically typical rather than favorable to concealment. To address this, we propose a \textbf{Contextual Reasoning Module (CRM)} using the knowledge encoded in a frozen MLLM to identify environments that are both semantically plausible and favorable to concealment, as shown in Fig.~\ref{fig:method}.

Environment reasoning without constraints can be unreliable because the general semantic knowledge of an MLLM is not aligned with the concealment objective, as illustrated in Fig.~\ref{fig:env}. The MLLM may select an ecologically plausible environment that leaves the target visually conspicuous (Fig.~\ref{fig:env}, Left), or it may produce an inaccurate environment prediction (Fig.~\ref{fig:env}, Right). We therefore restrict the reasoning space to a predefined candidate environment set $\mathcal{E}_{\mathrm{env}}$.

The environment set $\mathcal{E}_{\mathrm{env}}$ contains $100$ environment categories manually curated from Places365~\cite{zhou2017places} and publicly available web sources, covering diverse natural and built environments. For each input foreground object, CRM follows two successive stages: semantic candidate proposal and compatibility ranking. The MLLM proposes candidate environments using the object category and a concise description of its appearance, while a frozen CLIP model~\cite{radford2021learning} ranks these candidates according to their compatibility with the foreground. Together, these stages determine a coarse semantic context, whereas IAM subsequently regulates color and texture relations during background generation within the diffusion model.

Specifically, we use a frozen, quantized LLaVA-1.6 model~\cite{liu2024llavanext,liu2023improvedllava,liu2023llava} to propose a subset of candidate environments from $\mathcal{E}_{\mathrm{env}}$. Given the foreground $\mathbf X_{\mathrm{fg}}$, the MLLM predicts its object category $o$ and a concise visual description $d$, and selects $K$ candidate environments from $\mathcal{E}_{\mathrm{env}}$ to form $\mathcal{E}_{\mathrm{rec}}$, without introducing categories outside the predefined set. We then use the frozen CLIP model to rank the candidates according to their compatibility with the foreground. Specifically, the foreground image $\mathbf X_{\mathrm{fg}}$ and each candidate environment label $e_i \in \mathcal{E}_{\mathrm{rec}}$ are encoded into the shared CLIP embedding space using the image encoder $f_{\mathrm{img}}$ and the text encoder $f_{\mathrm{txt}}$, respectively. The candidate with the highest cosine similarity is selected as $e^*$:
\begin{equation}
e^*
=
\operatorname*{argmax}_{e_i \in \mathcal{E}_{\mathrm{rec}}}
\operatorname{sim}
\left(
f_{\mathrm{img}}(\mathbf X_{\mathrm{fg}}),
f_{\mathrm{txt}}(e_i)
\right),
\end{equation}
where $\operatorname{sim}(\cdot,\cdot)$ denotes cosine similarity.

Finally, CRM constructs a semantic prompt using the template
\emph{``A photo of an [$o$] with [$d$], camouflaged in [$e^*$].''}
The semantic condition $\mathbf C_{\mathrm{sem}}$ is obtained by encoding this prompt with the native text encoder configuration of the selected diffusion backbone at inference time.

\begin{figure}[!t]
\centering
\includegraphics[width=0.98\linewidth]{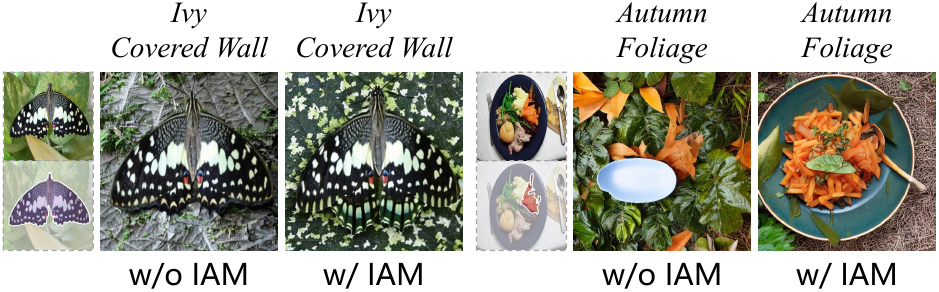}
\caption{Visual comparison of images generated with and without IAM. The text above each image denotes the inferred environment.}
\label{fig:ism}
\end{figure}

\subsection{Intrinsic Appearance Module}
\label{sec:ism}

Although CRM provides a semantic condition $\mathbf C_{\mathrm{sem}}$ that identifies an environment favorable to concealment, it captures only coarse contextual information and does not encode appearance information specific to the foreground instance, such as dominant color and texture patterns that are critical for concealment. As illustrated in Fig.~\ref{fig:ism}, even when CRM selects a suitable environment, the generated background may remain inconsistent with the foreground in color and texture, particularly near the object boundary. For example, the generated \textit{Autumn Foliage} fails to align with the orange appearance of the carrot, leaving the object visually conspicuous. Similarly, the background generated for the butterfly lacks the local high-frequency patterns required for texture continuity.

To address this limitation, we introduce the \textbf{Intrinsic Appearance Module (IAM)} to complement semantic compatibility with appearance assimilation. IAM extracts appearance cues from the foreground and injects them into the generation process, guiding the synthesized background toward greater consistency with the target in color and texture. Specifically, as illustrated in Fig.~\ref{fig:method}, IAM uses two complementary branches to extract \emph{chromatic} and \emph{textural} cues from the foreground.

The chromatic branch captures the dominant color distribution and illumination characteristics of the foreground while suppressing fine texture details. Given the foreground image $\mathbf X_{\mathrm{fg}}$, we first transform it into the LAB color space to separate luminance from chromatic information. An intensity aggregation operator is then applied to normalize the chromatic statistics with respect to foreground luminance, followed by Gaussian blurring to suppress fine texture patterns. The resulting chromatic representation is encoded as
\begin{equation}
\mathbf F_{\mathrm{chr}}
=
f_{\mathrm{img}}^{(8)}
\left(
\mathcal G
\left(
\Psi_{\mathrm{agg}}
\left(
\mathrm{LAB}(\mathbf X_{\mathrm{fg}})
\right)
\right)
\right),
\label{eq:chromatic_feature}
\end{equation}
where $\mathcal G$ denotes Gaussian blurring and $\Psi_{\mathrm{agg}}(\cdot)$ denotes the intensity aggregation operator. The function $f_{\mathrm{img}}^{(8)}(\cdot)$ represents the feature embedding extracted from the eighth block of the frozen CLIP image encoder. This intermediate representation retains low-level appearance details while preserving sufficient semantic structure. An analysis of the selected feature layer is provided in Fig.~\ref{fig:clip_layer}.

The texture branch extracts high-frequency cues that describe local surface patterns and details near object boundaries, both of which influence texture continuity between the target and its surroundings. We first convert $\mathbf X_{\mathrm{fg}}$ to grayscale to remove chromatic information. Histogram equalization (HE) is then applied to enhance local contrast, followed by a high-pass filter (HPF) that suppresses low-frequency components. The resulting texture representation is encoded as
\begin{equation}
\mathbf F_{\mathrm{tex}}
=
f_{\mathrm{img}}^{(8)}
\left(
\mathrm{HPF}
\left(
\mathrm{HE}
\left(
\mathrm{Gray}(\mathbf X_{\mathrm{fg}})
\right)
\right)
\right).
\label{eq:texture_feature}
\end{equation}

Finally, the appearance condition $\mathbf C_{\mathrm{app}}$ is obtained by fusing $\mathbf F_{\mathrm{chr}}$ and $\mathbf F_{\mathrm{tex}}$ through element-wise addition:
\begin{equation}
\mathbf C_{\mathrm{app}}
=
\mathbf F_{\mathrm{chr}}
+
\mathbf F_{\mathrm{tex}}.
\label{eq:appearance_condition}
\end{equation}
Using the token concatenation mechanism in Eq.~\ref{eq:fusion}, $\mathbf C_{\mathrm{app}}$ is injected only into UpBlock-2 starting from the 10th denoising step. These choices avoid over-constraining high-resolution generation while allowing IAM to refine local color and texture after the coarse scene layout has formed, as analyzed in Fig.~\ref{fig:abl}(b) and (d).

% The condition $\mathbf C_{\mathrm{app}}$ is injected into the frozen U-Net decoder at the second Upsampling Block (UpBlock-2). Notably, $\mathbf C_{\mathrm{app}}$ is \emph{not} applied from the beginning of the denoising process. Instead, $\mathbf C_{\mathrm{app}}$ is injected starting from the 10th denoising step and disabled otherwise.

\begin{table*}[!t]
\centering
\caption{Quantitative comparison on LAKE-RED using FID and KID across the camouflaged objects, salient objects, general objects, and overall subsets. Lower scores indicate better generation quality. \textbf{Bold} and \underline{underlined} values indicate the best and second-best results in each column, respectively.}
\label{tab:sota_comparison}

\renewcommand{\arraystretch}{1.06}
\resizebox{\linewidth}{!}{%
\begin{tabular}{l@{\hskip 1pt}cc |cc|cc|cc|cc}
\toprule
\rowcolor{gray!10}
&  &  & \multicolumn{2}{c|}{\hspace{-3pt} Camouflaged Objects} & \multicolumn{2}{c|}{Salient Objects} & \multicolumn{2}{c|}{General Objects} & \multicolumn{2}{c}{Overall} \\
\arrayrulecolor{gray!10}
\midrule
\arrayrulecolor{black}
\rowcolor{gray!10}                          
\multirow{-2}{*}{Method} & \multirow{-2}{*}{Paradigm} & \multirow{-2}{*}{Training} & \hspace{2pt}
FID $\downarrow$ & \hspace{2pt}  KID $\downarrow$ & \hspace{2pt}
FID $\downarrow$ & \hspace{2pt}  KID $\downarrow$ & \hspace{2pt}
FID $\downarrow$ & \hspace{2pt}  KID $\downarrow$ & \hspace{2pt}
FID $\downarrow$ & \hspace{2pt}  KID $\downarrow$ \\ \midrule \midrule
% AB~\cite{AB}\textcolor[HTML]{9B9B9B}{\scriptsize[{IPOL}]} & Classical & No & 117.11 & 0.0645 & 126.78 & 0.0614 & 133.89 & 0.0645 & 120.21 & 0.0623 \\
% CI~\cite{CI}\textcolor[HTML]{9B9B9B}{\scriptsize[{J. Zool.}]} & Classical & No & 124.49 & 0.0662 & 136.30 & 0.0738 & 137.19 & 0.0713 & 128.51 & 0.0693 \\
% AdaIN~\cite{AdaIN}\textcolor[HTML]{9B9B9B}{\scriptsize[{ICCV}]} & CNN & Yes & 125.16 & 0.0721 & 133.20 & 0.0702 & 136.93 & 0.0714 & 126.94 & 0.0703 \\
% DCI~\cite{DCI}\textcolor[HTML]{9B9B9B}{\scriptsize[{AAAI}]} & CNN & Yes & 130.21 & 0.0689 & 134.92 & 0.0665 & 137.99 & 0.0690 & 130.52 & 0.0673 \\
LCGNet~\cite{LCGNet}\textcolor[HTML]{9B9B9B}{\scriptsize[{TMM}]} & CNN & Yes & 129.80 & 0.0504 & 136.24 & 0.0597 & 132.64 & 0.0548 & 129.88 & 0.0550 \\ 
TFill~\cite{TFill}\textcolor[HTML]{9B9B9B}{\scriptsize[{CVPR}]} & CNN & Yes & 63.74 & 0.0336 & 96.91 & 0.0453 & 122.44 & 0.0747 & 80.39 & 0.0438 \\
LDM~\cite{LDM}\textcolor[HTML]{9B9B9B}{\scriptsize[{CVPR}]} & Diffusion & Yes & 58.65 & 0.0380 & 107.38 & 0.0524 & 129.04 & 0.0748 & 84.48 & 0.0488 \\
LAKE-RED~\cite{zhao2024lake}\textcolor[HTML]{9B9B9B}{\scriptsize[{CVPR}]} & Diffusion & Yes & 39.55 & 0.0212 & 88.70 & 0.0428 & 102.67 & 0.0555 & 64.27 & 0.0355 \\
FACIG~\cite{chen2025foreground}\textcolor[HTML]{9B9B9B}{\scriptsize[{ICME}]} & Diffusion & Yes & 27.61 & 0.0099 & 82.23 & 0.0326 & 96.94 & 0.0503 & 52.87 & 0.0229 \\
CamAny~\cite{das2025camouflage}\textcolor[HTML]{9B9B9B}{\scriptsize[{CVPR}]} & Diffusion & Yes & \underline{22.30} & \underline{0.0039} & 61.78 & 0.0211 & 74.53 & 0.0387 & 40.53 & 0.0155 \\ 
CT-CIG~\cite{CTCIG}\textcolor[HTML]{9B9B9B}{\scriptsize[{AAAI}]} & Diffusion & Yes & 30.59 & 0.0085 & 81.60 & 0.0230 & 104.46 & 0.0241 & 52.88 & 0.0169 \\
\midrule \midrule
\rowcolor{best}
FreeCam$_{\mathrm{SD1.5}}$ & Diffusion & No & 26.43 & 0.0047 & 43.75 & 0.0097 & \underline{44.38} & \textbf{0.0075} & 28.82 & \underline{0.0069} \\ 
\rowcolor{best}
FreeCam$_{\mathrm{SD2.0}}$ & Diffusion & No & 23.48 & 0.0045 & \underline{31.71} & \textbf{0.0079} & 54.77 & 0.0167 & \underline{26.91} & 0.0073 \\ 
\rowcolor{best}
FreeCam$_{\mathrm{SDXL}}$  & Diffusion & No & \textbf{22.16} & \textbf{0.0032} & \textbf{28.98} & \underline{0.0082} & \textbf{42.22} & \underline{0.0113} & \textbf{22.73} & \textbf{0.0065} \\ 

\bottomrule
\end{tabular}%
}
\end{table*}

\begin{table*}[!t]

\centering
\caption{Quantitative comparison of camouflage effectiveness on the LAKE-RED dataset using $S_{\mathrm{Rf}}$ and $S_{\mathrm{b}}$. Since these two metrics are not reported in the original papers of the compared CIG methods, all results are re-evaluated by us. $\uparrow$ indicates higher is better. \textbf{Bold} and \underline{underlined} values indicate the best and second-best results in each column, respectively.}
\label{tab:sota_similarity_comparison}
\resizebox{\linewidth}{!}{%
\begin{tabular}{l@{\hskip 1pt}cc |cc|cc|cc|cc}
\toprule
\rowcolor{gray!10}
&  &  & \multicolumn{2}{c|}{\hspace{-3pt} Camouflaged Objects} & \multicolumn{2}{c|}{Salient Objects} & \multicolumn{2}{c|}{General Objects} & \multicolumn{2}{c}{Overall} \\
\arrayrulecolor{gray!10}
\midrule
\arrayrulecolor{black}
\rowcolor{gray!10}                          
\multirow{-2}{*}{Method} & \multirow{-2}{*}{Paradigm} & \multirow{-2}{*}{Training} & \hspace{2pt}
$S_{\mathrm{Rf}}$  $\uparrow$ & \hspace{2pt}  $S_{\mathrm{b}}$ $\uparrow$ & \hspace{2pt}
$S_{\mathrm{Rf}}$ $\uparrow$ & \hspace{2pt} $S_{\mathrm{b}}$ $\uparrow$ & \hspace{2pt}
$S_{\mathrm{Rf}}$ $\uparrow$ & \hspace{2pt} $S_{\mathrm{b}}$ $\uparrow$ & \hspace{2pt}
$S_{\mathrm{Rf}}$ $\uparrow$ & \hspace{2pt} $S_{\mathrm{b}}$ $\uparrow$ \\ \midrule \midrule

LCGNet~\cite{LCGNet}\textcolor[HTML]{9B9B9B}{\scriptsize[{TMM}]} & CNN & Yes & 0.2724 & 0.4223 & 0.2845 & 0.3923 &0.2212  &0.4161  &0.2618  & 0.4030 \\ 
TFill~\cite{TFill}\textcolor[HTML]{9B9B9B}{\scriptsize[{CVPR}]} & CNN & Yes & 0.2862 & 0.2771 & 0.2604 & 0.2968 & 0.2271 & 0.3130 & 0.2273 & 0.3164 \\
LDM~\cite{LDM}\textcolor[HTML]{9B9B9B}{\scriptsize[{CVPR}]} & Diffusion & Yes & 0.3361 & 0.3480 & 0.4012 & 0.4539 & 0.3692 & 0.4317 & 0.3804 & 0.4576 \\
LAKE-RED~\cite{zhao2024lake}\textcolor[HTML]{9B9B9B}{\scriptsize[{CVPR}]} & Diffusion & Yes & 0.3715 & 0.4208 & 0.3941 & 0.4764 & 0.3209 & 0.4621 & 0.3533 & 0.4436 \\
FACIG~\cite{chen2025foreground}\textcolor[HTML]{9B9B9B}{\scriptsize[{ICME}]} & Diffusion & Yes & \underline{0.5725} & 0.4515 & 0.3992 & 0.4843 & 0.3575 & 0.4520 & 0.4057 & 0.4586 \\
CamAny~\cite{das2025camouflage}\textcolor[HTML]{9B9B9B}{\scriptsize[{CVPR}]} & Diffusion & Yes & 0.4780 & 0.4329 & 0.4074 & 0.4430 & 0.4611 & 0.4522  & 0.4329 & 0.4206 \\
CT-CIG~\cite{CTCIG}\textcolor[HTML]{9B9B9B}{\scriptsize[{AAAI}]} & Diffusion & Yes & 0.5240 & 0.6256 & 0.4863 & 0.5674 & 0.4718 & 0.6206 & 0.5085 & 0.6260 \\
\midrule \midrule
\rowcolor{best}
FreeCam$_{\mathrm{SD1.5}}$ & Diffusion & No & 0.5540 & \underline{0.8190} & \underline{0.5532} & 0.8103 & 0.5275 & 0.7794 & 0.5449 & 0.8029 \\
\rowcolor{best}
FreeCam$_{\mathrm{SD2.0}}$ & Diffusion & No & 0.5412 & 0.8166 & 0.5262 & \textbf{0.8387} & \textbf{0.5977} & \underline{0.7989} & \underline{0.5617} & \underline{0.8114} \\
\rowcolor{best}
FreeCam$_{\mathrm{SDXL}}$  & Diffusion & No & \textbf{0.6053} & \textbf{0.8201} & \textbf{0.5852} & \underline{0.8283} & \underline{0.5923} & \textbf{0.8359} & \textbf{0.5976} & \textbf{0.8347} \\

\bottomrule
\end{tabular}%
}
\end{table*}

\section{Experiments}

We organize the experiments around the central question of whether training-free CIG is viable and how FreeCam performs under this paradigm. First, we compare FreeCam with representative training-based CIG methods and generic generation baselines in terms of generation quality, camouflage effectiveness, and computational efficiency. Second, we conduct controlled ablations to examine how CRM and IAM address semantic compatibility and appearance assimilation, together with the associated conditioning designs. Foreground preservation is held fixed across all variants through the background inpainting formulation.

We then evaluate, from a downstream perspective, the utility of the generated images as synthetic training data for camouflaged object detection and examine whether they reduce target detectability when evaluated with general object detectors. Finally, we analyze controllability, generalization across scenarios and generative backbones, sampling stability, and representative failure cases.

\subsection{Experimental Setup}

\begin{figure*}[!t]
\centering
\includegraphics[width=0.98\linewidth]{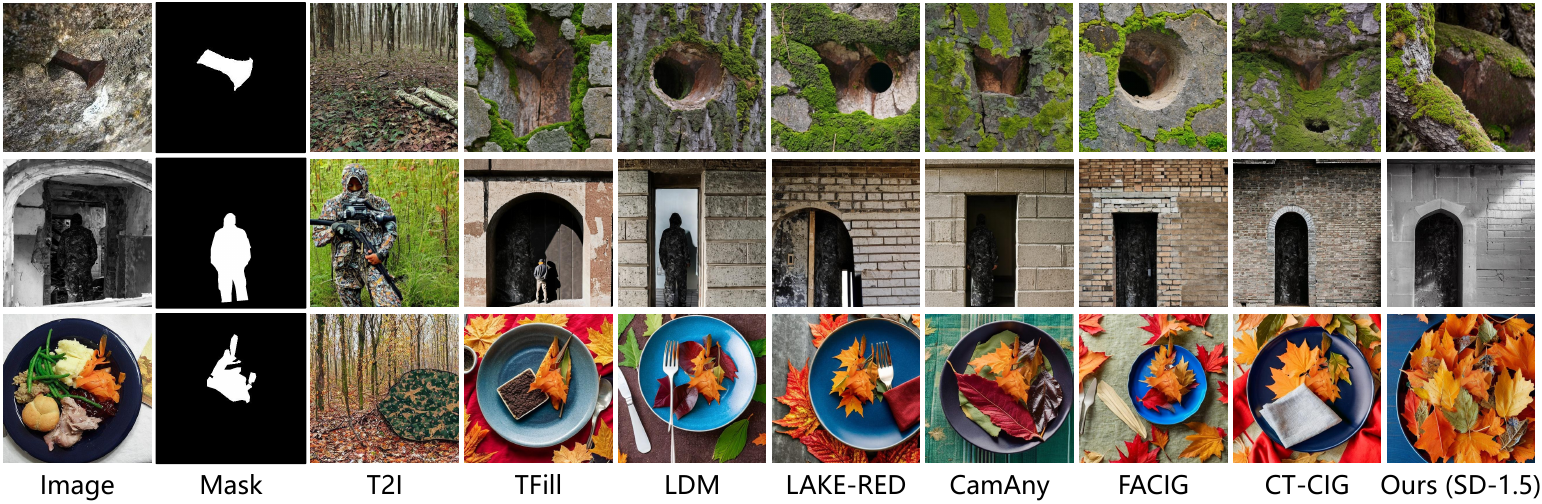}
\caption{Qualitative comparison with representative CIG methods.
The first two columns show the input images and masks.} 
\label{fig:results}
\end{figure*}

\noindent \textbf{Datasets.}
We conduct the main CIG evaluation on the LAKE-RED benchmark~\cite{zhao2024lake}, which has been adopted by several recent CIG studies~\cite{das2025camouflage,chen2025foreground,CTCIG}. The benchmark contains three subsets comprising camouflaged objects (CO), salient objects (SO), and general objects (GO). The CO subset aggregates concealed targets from CAMO~\cite{le2019anabranch}, COD10K~\cite{fan2020camouflaged}, and NC4K~\cite{lv2021simultaneously}. The SO subset contains visually prominent foregrounds from DUT-OMRON~\cite{yang2013saliency} and DUTS~\cite{wang2017learning}. The GO subset samples objects from MS-COCO~\cite{lin2014microsoft}. Each subset contains $6{,}473$ image--mask pairs. This protocol evaluates performance on CO, SO, and GO under consistent settings, with CO being closer to camouflage-specific training distributions and SO and GO providing a broader test of generalization through foreground categories and visual statistics.

\noindent \textbf{Baselines.}
Following prior CIG studies~\cite{zhao2024lake,das2025camouflage,chen2025foreground,CTCIG}, we compare FreeCam with representative CNN and diffusion methods, including LCGNet~\cite{LCGNet}, TFill~\cite{TFill}, LDM~\cite{LDM}, LAKE-RED~\cite{zhao2024lake}, FACIG~\cite{chen2025foreground}, CamAny~\cite{das2025camouflage}, and CT-CIG~\cite{CTCIG}. These baselines cover supervised appearance-transfer networks, general-purpose inpainting models, and camouflage-specific diffusion generators, enabling comparison across different modeling paradigms and levels of task-specific optimization within a unified experimental setting.

\begin{table}[!t]
\centering
\caption{Comparison of preprocessing time and inference time. Preprocessing time includes the time required for fine-tuning or re-training. \textbf{Bold} indicates the lowest total time.}
\label{tab:efficient}
\begingroup
\scriptsize
\setlength{\tabcolsep}{4.5pt}
\renewcommand{\arraystretch}{1.2}
\resizebox{0.90\linewidth}{!}{%
\begin{tabular}{lccc}
\toprule
\rowcolor{gray!10}
Method & Preprocessing & Inference & Total \\
\midrule
LCGNet~\cite{LCGNet} & 70 min & 0.05 s & 70 min \\
LAKE-RED~\cite{zhao2024lake} & 5 h & 9.8 s & 5 h \\
FACIG~\cite{chen2025foreground} & 6 h & 3.7 s & 6 h \\
CamAny~\cite{das2025camouflage} & $>$1 day & 13.2 s & $>$1 day \\
CT-CIG~\cite{CTCIG} & 8 h & 23.1 s & 8 h \\
\midrule
\rowcolor{best}
FreeCam$_{\mathrm{SD1.5}}$ & 0 s & 8.3 s & \textbf{8.3 s} \\
\rowcolor{best}
FreeCam$_{\mathrm{SD2.0}}$ & 0 s & 17.1 s & 17.1 s \\
\rowcolor{best}
FreeCam$_{\mathrm{SDXL}}$ & 0 s & 28.9 s & 28.9 s \\
\bottomrule
\end{tabular}%
}
\endgroup
% \vspace{-1.0em}
\end{table}

\noindent \textbf{Implementation Details.}
Unless otherwise specified, FreeCam uses \textit{Stable Diffusion~1.5 Inpainting}~\cite{LDM} as the default frozen backbone. To evaluate its compatibility with different generative backbones, we also implement FreeCam with \textit{Stable Diffusion~2.0 Inpainting}~\cite{LDM} and \textit{SDXL Inpainting}~\cite{sdxl}. The three variants are denoted as FreeCam$_{\mathrm{SD1.5}}$, FreeCam$_{\mathrm{SD2.0}}$, and FreeCam$_{\mathrm{SDXL}}$, respectively. All model components remain frozen during inference.
In CRM, a quantized LLaVA-1.6 model~\cite{liu2024llavanext} selects $K=5$ candidate environments from a predefined set of 100 scene categories. A frozen CLIP model then ranks these candidates according to their compatibility with the foreground object. In IAM, the foreground appearance condition is extracted from the eighth transformer block of CLIP ViT-B/16~\cite{radford2021learning}. The same CLIP model is used for candidate ranking in CRM. The appearance condition is fused with the semantic condition through the frozen cross-attention layers within the diffusion model and injected into UpBlock-2 starting from the 10th denoising step.

All images and masks are resized to $512\times512$. We use DDIM~\cite{song2020denoising} with 50 denoising steps for sampling. For the main generation metrics, we repeat the full generation process with five random seeds and report the mean score across the five runs. The effects of $K$, the number of scene categories, the injection block, and the injection start step are evaluated in the ablation studies. All experiments are implemented with the Diffusers library~\cite{diffuser} and conducted on a single NVIDIA A100 GPU.

\begin{table*}[!t]
\centering
\caption{Main ablation of FreeCam. The table summarizes the route from a frozen prior to FreeCam by isolating CRM and IAM. KID is scaled by $10^{3}$. Lower KID and higher $S_{\mathrm{b}}$ are better. \textbf{Bold} indicates the best value for each metric.}
\label{tab:main_ablation_hybrid}
\begingroup
\setlength{\tabcolsep}{6.0pt}
\renewcommand{\arraystretch}{1.1}
\newcommand{\cmark}{\ding{51}}
\newcommand{\xmark}{\ding{55}}
\resizebox{0.6\textwidth}{!}{%
\begin{tabular}{ccccc}
\toprule
\rowcolor{gray!10}
CRM & IAM-Color & IAM-Texture & KID $\downarrow$ & $S_{\mathrm{b}}\uparrow$ \\
\midrule
% \rowcolor{gray!5}
% \multicolumn{5}{c}{\textit{Semantic-condition construction (IAM off)}} \\
\xmark   & \xmark & \xmark & 13.326 & 0.2421 \\
% MLLM  & \xmark & \xmark & 9.643  & 0.5304 \\
\midrule \rowcolor{gray!5}
\multicolumn{5}{c}{\textit{IAM-based ablation without CRM}} \\
\xmark & \cmark & \xmark & 8.981  & 0.7096 \\
\xmark & \xmark & \cmark & 11.424 & 0.6818 \\
\xmark & \cmark & \cmark & 8.655 & 0.7631 \\
\midrule \rowcolor{gray!5}
\multicolumn{5}{c}{\textit{CRM-based ablation}} \\
% \midrule
Manual Prompt & \xmark & \xmark & 12.219 & 0.4397 \\
CRM w/o Env.Set & \xmark & \xmark & 8.161 & 0.5953 \\
\cmark  & \xmark & \xmark & 8.244  & 0.6439 \\
\cmark  & \cmark & \xmark & 7.016  & 0.7561 \\
\cmark  & \xmark & \cmark & 7.440  & 0.7371 \\
\midrule \rowcolor{best}
\cmark  & \cmark & \cmark & \textbf{6.926} & \textbf{0.8029} \\
\bottomrule
\end{tabular}%
}
\endgroup
\end{table*}

\begin{figure*}[!t]
\centering
\includegraphics[width=0.85\linewidth]{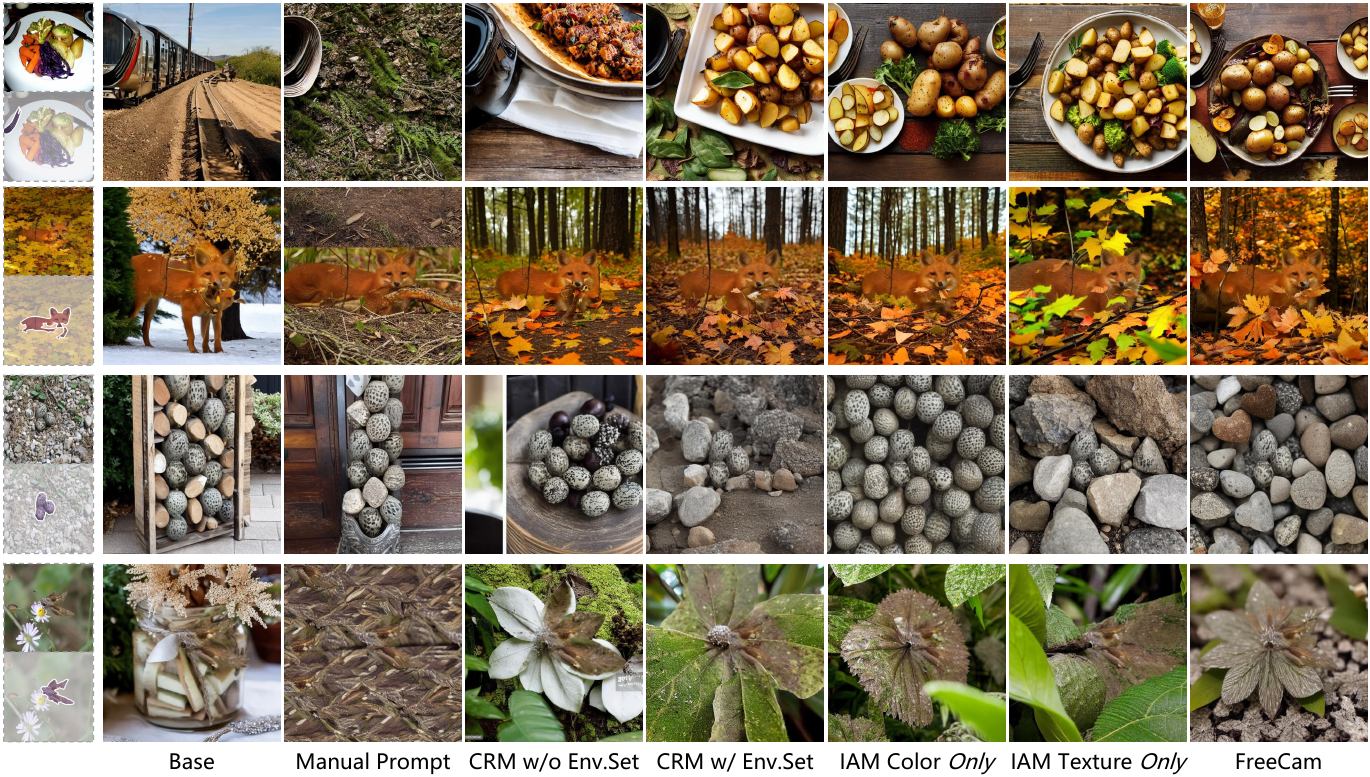}
\caption{Qualitative component ablation showing the progression from the frozen baseline to the complete FreeCam framework.}
\label{fig:base_to_full}
\end{figure*}

\noindent \textbf{Metrics.}
We evaluate FreeCam from three perspectives. \emph{(i) Generation quality.} We report Fr\'echet Inception Distance (FID)~\cite{FID} and Kernel Inception Distance (KID)~\cite{KID}, computed using InceptionV3 features~\cite{inception}. Lower values indicate smaller distributional discrepancies between synthesized and real images. \emph{(ii) Camouflage effectiveness.} Since distributional metrics do not directly measure whether the target is concealed, we further adopt two complementary scores from~\cite{Lamdouar_2023_ICCV}. The reconstruction fidelity $S_{\mathrm{Rf}}$ measures how well foreground appearance can be reconstructed from nearby background patches and provides an indirect measure of appearance compatibility. The boundary score $S_{\mathrm{b}}$ evaluates the visibility of the object contour within a boundary band, with a higher value indicating a less distinguishable boundary. \emph{(iii) Downstream perception.} For camouflaged object detection, we report the Structure measure ($S_{\alpha}$), Enhanced-alignment measure ($E_{\phi}$), weighted F-measure ($F_{\beta}^{w}$), and Mean Absolute Error ($M$), where higher $S_{\alpha}$, $E_{\phi}$, and $F_{\beta}^{w}$ and lower $M$ indicate better detection performance. For the YOLOv10-X and GroundingDINO-B (G-DINO-B) detectors, we report Average Precision (AP) and recall at IoU thresholds 0.50/0.75, mean Average Precision (mAP) over IoU thresholds 0.50:0.05:0.95, and intersection over union (IoU).

\subsection{Main Results}

\begin{table*}[!t]
\centering
\caption{Ablation analysis of CRM. (a) evaluates prompt composition, and (b) compares environment-selection strategies in the CRM scorer. KID is scaled by $10^{3}$. Lower KID and higher $S_{\mathrm{b}}$ are better. \textbf{Bold} indicates the best value in each subtable.}
\label{tab:abl_crm}
\begin{minipage}{0.78\textwidth}
\begingroup
\footnotesize
\setlength{\tabcolsep}{1pt}
\renewcommand{\arraystretch}{1.05}
\newcommand{\abltitle}[1]{%
    \parbox[t][2.10\baselineskip][c]{\linewidth}{\centering\bfseries #1}\par\vspace{1pt}}
\newcommand{\kidhead}{KID\(\downarrow\)}
\newcommand{\sbhead}{$S_{\mathrm{b}}\uparrow$}

\begin{minipage}[t]{0.48\linewidth}
    \centering
    \abltitle{(a) Prompt composition}
    \begin{tabularx}{\linewidth}{@{}>{\raggedright\arraybackslash}X>{\centering\arraybackslash}p{3.5em}>{\centering\arraybackslash}p{3.5em}@{}}
    \toprule 
    \rowcolor{gray!10}
    Prompt & \kidhead & \sbhead \\ 
    \midrule
    Base (IAM)    & 8.655 & 0.7631\\
    +Category          & 8.135 & 0.7697\\
    +Description       & 7.315 & 0.7846\\
    +Environment w/o Env.Set   & \textbf{6.859} & 0.7959\\
    +Env.Set       & 6.926 & \textbf{0.8029}\\
    \bottomrule
    \end{tabularx}
\end{minipage}%
\hfill
\begin{minipage}[t]{0.48\linewidth}
    \centering
    \abltitle{(b) CLIP scorer strategy in CRM}
    \begin{tabularx}{\linewidth}{@{}>{\raggedright\arraybackslash}X>{\centering\arraybackslash}p{3.5em}>{\centering\arraybackslash}p{3.5em}@{}}
    \toprule
    \rowcolor{gray!10}
    Strategy & \kidhead & \sbhead \\
    \midrule
    Random from top-$K$ & 9.693 & 0.7867\\
    Top-$1$ by LLaVA-1.6~\cite{liu2024llavanext}          & 7.136 & 0.7953\\
    $\operatorname{sim}(f_{\mathrm{txt}}(o,d),f_{\mathrm{txt}}(e_i))$    & 8.005 & 0.7992\\
    $\operatorname{sim}(f_{\mathrm{img}}(\mathbf X_{\mathrm r}),f_{\mathrm{txt}}(e_i))$ & 7.131 & 0.8024\\
    $\operatorname{sim}(f_{\mathrm{img}}(\mathbf X_{\mathrm{fg}}),f_{\mathrm{txt}}(e_i))$& \textbf{6.926} & \textbf{0.8029}\\
    \bottomrule
    \end{tabularx}
\end{minipage}

\endgroup
\end{minipage}

\end{table*}

\begin{table*}[!t]
\centering
\caption{Ablation analysis of IAM. (a) compares filter choices for texture extraction, and (b) evaluates cumulative IAM components. KID is scaled by $10^{3}$. Lower KID and higher $S_{\mathrm{b}}$ are better. \textbf{Bold} indicates the best value in each subtable.}
\label{tab:abl_iam}
\begin{minipage}{0.78\textwidth}
\begingroup
\footnotesize
\setlength{\tabcolsep}{1pt}
\renewcommand{\arraystretch}{1.05}
\newcommand{\abltitle}[1]{%
    \parbox[t][2.10\baselineskip][c]{\linewidth}{\centering\bfseries #1}\par\vspace{1pt}}
\newcommand{\kidhead}{KID\(\downarrow\)}
\newcommand{\sbhead}{$S_{\mathrm{b}}\uparrow$}

\begin{minipage}[t]{0.48\linewidth}
    \centering 
    \abltitle{(a) Filter algorithms in IAM}
    \begin{tabularx}{\linewidth}{@{}>{\raggedright\arraybackslash}X>{\centering\arraybackslash}p{3.5em}>{\centering\arraybackslash}p{3.5em}@{}}
    \toprule
    \rowcolor{gray!10}
    Filter & \kidhead & \sbhead \\ 
    \midrule
    Sobel     & 7.099 & 0.7994\\
    Laplacian & 7.203 & 0.8011\\
    Canny     & 7.144 & 0.8002\\
    DoG       & 7.065 & 0.8026\\
    High-pass & \textbf{6.926} & \textbf{0.8029}\\ 
    \bottomrule
    \end{tabularx}
\end{minipage}%
\hfill
\begin{minipage}[t]{0.48\linewidth}
    \centering 
    \abltitle{(b) IAM component analysis}
    \begin{tabularx}{\linewidth}{@{}>{\raggedright\arraybackslash}X>{\centering\arraybackslash}p{3.5em}>{\centering\arraybackslash}p{3.5em}@{}}
    \toprule
    \rowcolor{gray!10}
    Component & \kidhead & \sbhead \\
    \midrule
    Base (CRM)    & 8.244 & 0.6439\\
    +LAB    & 7.852 & 0.7812\\
    +Blur       & 7.535 & 0.7901\\
    +HE        & 7.410 & 0.7889\\
    +High-pass & \textbf{6.926} & \textbf{0.8029}\\
    \bottomrule
    \end{tabularx}
\end{minipage}

\endgroup
\end{minipage}

\end{table*}

\noindent \textbf{Generation quality.}
Tab.~\ref{tab:sota_comparison} compares generation quality on LAKE-RED using FID and KID. The new FreeCam method achieves the lowest FID on all four subsets and the lowest KID on CO, SO, and Overall without optimization on CODs; on GO KID, FreeCam$_{\mathrm{SD1.5}}$ remains the best and the new method is second-best. The new method reduces the overall FID from $40.53$ for the strongest training-based baseline to $22.73$, with KID decreasing from $0.0155$ to $0.0065$. These gains show that pretrained generative priors can be effectively redirected toward camouflage synthesis. On CO, which is closer to existing CIG training distributions, the new method surpasses CamAny, while its advantage is also maintained on SO and GO. This contrast suggests that FreeCam benefits primarily from preserving the broad domain coverage of the pretrained diffusion prior rather than increasing specialization to camouflage data. The variation across backbones further indicates that performance depends on the compatibility between foreground guidance and the visual knowledge encoded by each prior. Nevertheless, the strongest overall performance of the new method shows that training-free CIG can benefit from stronger generative backbones without optimization on CODs.

\noindent \textbf{Camouflage effectiveness.}
Tab.~\ref{tab:sota_similarity_comparison} evaluates whether the generated backgrounds reduce foreground separability. FreeCam$_{\mathrm{SDXL}}$ achieves the best overall $S_{\mathrm{Rf}}$ and $S_{\mathrm{b}}$, improving them from $0.5085$ and $0.6260$ for the strongest training-based competitor to $0.5976$ and $0.8347$, respectively. The advantage is especially consistent for $S_{\mathrm{b}}$, with all FreeCam variants producing substantially less visible boundaries across the three subsets. The higher $S_{\mathrm{Rf}}$ also suggests stronger local appearance compatibility between the target and its surroundings. This joint pattern, observed at both local and boundary levels, reflects the complementary roles of CRM and IAM, with CRM selecting contexts favorable to concealment and IAM regulating local appearance relations.

\begin{figure*}[!t]
    \centering
    \includegraphics[width=0.98\linewidth]{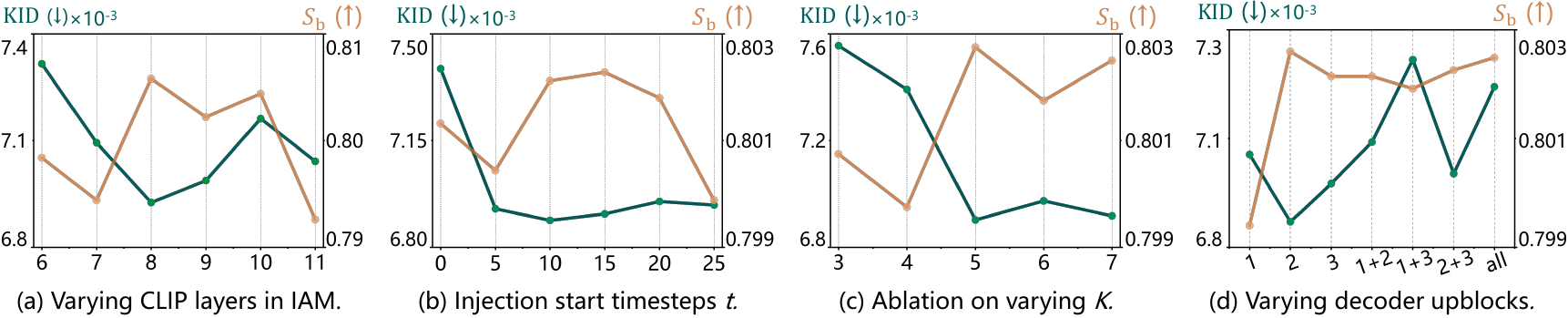}
    \caption{Sensitivity analysis of hyperparameters and architectural choices in terms of KID ($\downarrow$) and $S_{\mathrm{b}}$ ($\uparrow$).}
    \label{fig:abl}
\end{figure*}

\begin{figure*}[!t]
\centering
\includegraphics[width=0.8\linewidth]{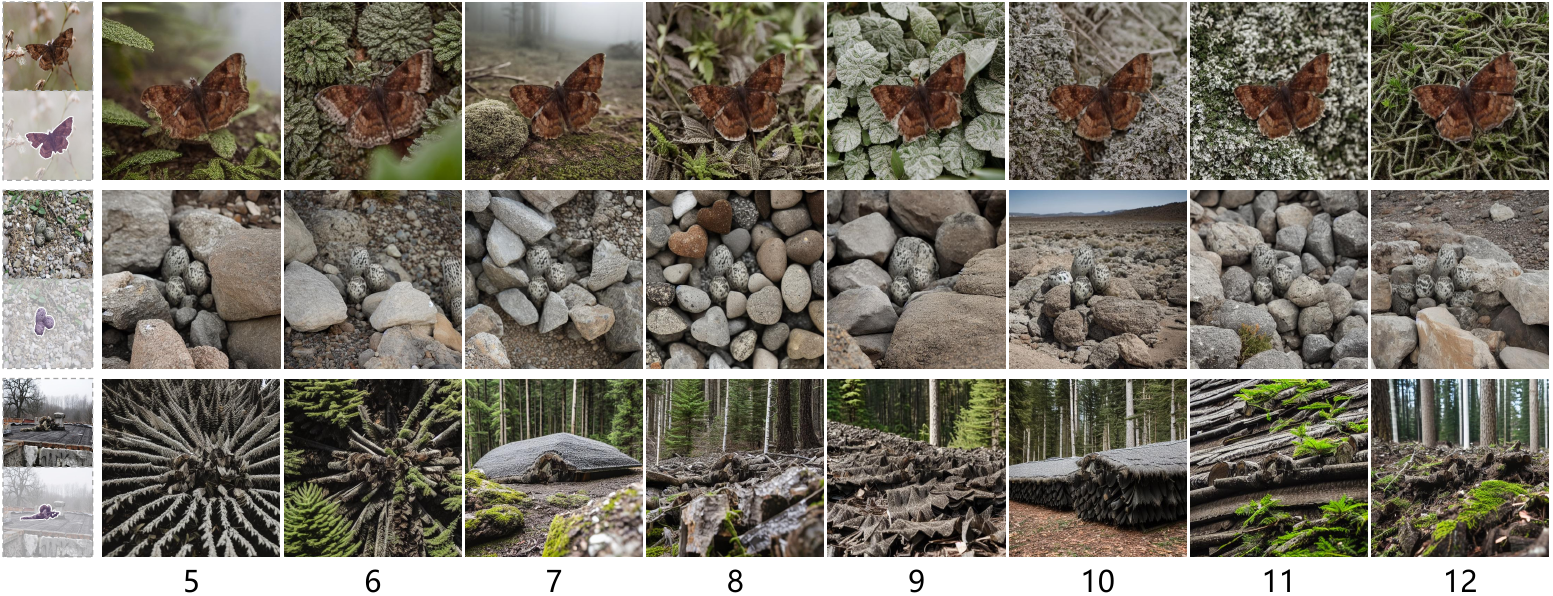}
\caption{Effect of the CLIP feature layer used for intrinsic appearance extraction. Intermediate features provide the best balance between texture preservation and semantic abstraction. The labels 5--12 denote the corresponding layers of CLIP.}
\label{fig:clip_layer}
\end{figure*}

\noindent \textbf{Efficiency and qualitative comparison.}
Tab.~\ref{tab:efficient} compares the time required to produce the first test image, including method-specific training or preparation. Training-based CIG methods require from $70$ minutes to more than one day before inference, whereas FreeCam$_{\mathrm{SD1.5}}$ produces its first result in $8.3$ seconds without parameter optimization for the target task. This advantage concerns adaptation and deployment cost rather than per-image throughput: LCGNet remains faster after training, and its initial cost may be amortized in a fixed large-scale workload. FreeCam is therefore better suited to changing deployment conditions that would otherwise require renewed optimization. Fig.~\ref{fig:results} further shows that CNN methods tend to suppress local texture details, while trained diffusion methods may produce realistic backgrounds that remain incompatible with the target in local appearance or scene context. By separating context selection from appearance regulation, FreeCam redirects the scene synthesis capability of the frozen prior toward concealment.

\subsection{Ablation Studies}
We organize the ablation studies around two camouflage-specific challenges and their implementation in FreeCam. First, we systematically trace the progression from a frozen inpainting prior to the complete camouflage generator. Second, we assess the key CRM and IAM designs for semantic compatibility and appearance assimilation. Third, we analyze sensitivity to the conditioning schedule, environment search space, and underlying foundation components. Unless stated otherwise, all experiments use the SD1.5 backbone, report overall KID scaled by $10^{3}$, and adopt $S_{\mathrm{b}}$ as the boundary camouflage metric.

\noindent \textbf{Ablation of the main design path.}
Tab.~\ref{tab:main_ablation_hybrid} and Fig.~\ref{fig:base_to_full} systematically trace the complete progression from a frozen inpainting prior to the complete FreeCam framework. The frozen baseline already preserves the foreground, yet its poor KID and $S_{\mathrm{b}}$ values show that preservation alone is insufficient for effective camouflage synthesis. Manual prompting reduces part of the semantic mismatch, but the selected context is specified by the user rather than automatically inferred from the foreground. CRM produces a larger overall improvement in KID, indicating that environment reasoning conditioned on the target mainly strengthens semantic compatibility. In contrast, IAM without CRM yields a much larger gain in $S_{\mathrm{b}}$, showing that foreground color and texture cues more directly improve local concealment even when context selection remains imperfect. Neither component is sufficient alone: CRM lacks appearance regulation for the individual target, whereas IAM cannot determine an environment favorable to concealment. Their combination consistently achieves the best overall KID and $S_{\mathrm{b}}$, demonstrating that context selection and local appearance assimilation address complementary sources of foreground separability.

% Tab.~\ref{tab:main_ablation_hybrid} and Fig.~\ref{fig:base_to_full} trace the progression from a frozen inpainting prior to the complete FreeCam framework. The frozen baseline already preserves the foreground, yet its poor KID and $S_{\mathrm{b}}$ show that preservation alone is insufficient for camouflage synthesis. Manual prompting reduces part of the semantic mismatch, but the selected context is specified by the user rather than inferred from the foreground. CRM produces a larger improvement in KID, indicating that environment reasoning conditioned on the target mainly strengthens semantic compatibility. In contrast, IAM without CRM yields a much larger gain in $S_{\mathrm{b}}$, showing that foreground color and texture cues directly improve local concealment even when context selection remains imperfect. Neither component is sufficient alone: CRM lacks appearance regulation for the individual target, whereas IAM cannot determine an environment favorable to concealment. Their combination achieves the best KID and $S_{\mathrm{b}}$, demonstrating that context selection and local appearance assimilation address complementary sources of foreground separability.

\noindent \textbf{Effectiveness of the Contextual Reasoning Module.}
Tab.~\ref{tab:abl_crm} validates the two components of CRM. Adding the object category and visual description improves both KID and $S_{\mathrm{b}}$, indicating that context selection benefits from semantic cues grounded in the foreground. Although removing the environment set slightly improves KID, it reduces $S_{\mathrm{b}}$, showing that distributional realism alone does not ensure compatibility with concealment. The scorer comparison favors the isolated foreground over text-only matching or the reference image, since it retains target-specific information while avoiding irrelevant background cues.

\begin{figure*}[!t]
\centering
\includegraphics[width=0.6\linewidth]{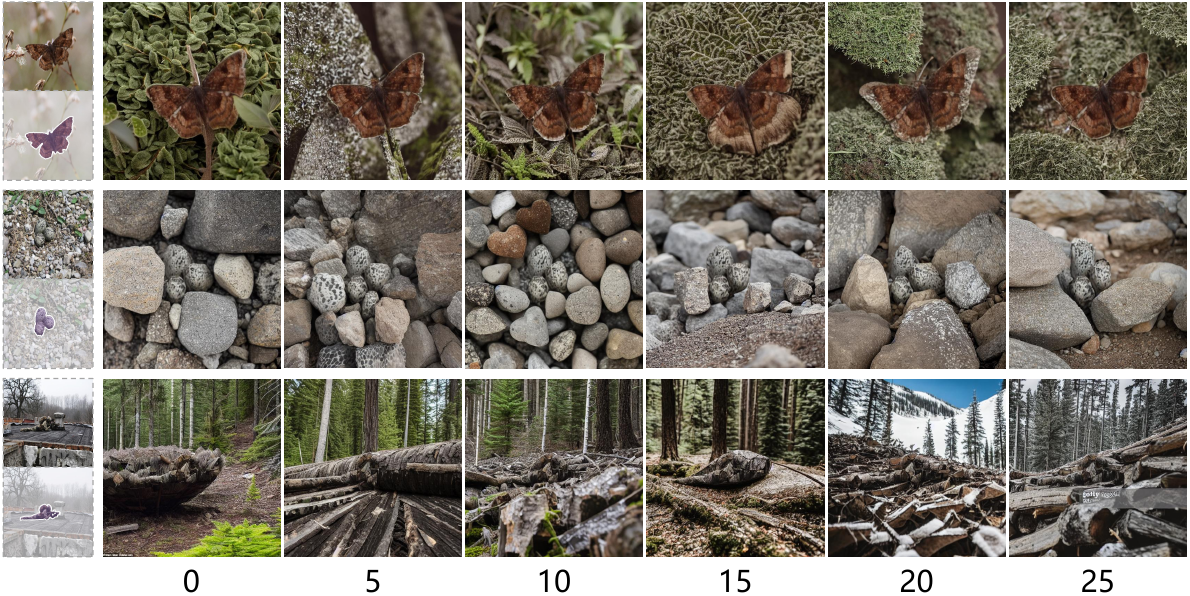}
\caption{Fine-grained analysis of the starting timestep for intrinsic appearance injection. Starting at an intermediate denoising stage improves appearance consistency while preserving the global scene layout. The labels denote the injection timesteps.}
% \caption{Fine-grained analysis of the starting timestep for intrinsic appearance injection. The comparison shows that appearance guidance should be introduced at an appropriate diffusion stage to improve texture continuity without over-constraining the generative process.}
\label{fig:start_t}
\end{figure*}

\begin{table*}[!t]
\centering
\caption{Model-choice sensitivity of FreeCam. (a) evaluates vision encoders used by IAM, and (b) evaluates MLLMs used by CRM. KID is scaled by $10^{3}$. Lower KID and higher $S_{\mathrm{b}}$ are better. \textbf{Bold} indicates the best value in each subtable.}
\label{tab:abl_model_choice}
\begin{minipage}{0.78\textwidth}
\begingroup
\footnotesize
\setlength{\tabcolsep}{1pt}
\renewcommand{\arraystretch}{1.05}
\newcommand{\abltitle}[1]{%
    \parbox[t][2.10\baselineskip][c]{\linewidth}{\centering\bfseries #1}\par\vspace{1pt}}
\newcommand{\kidhead}{KID\(\downarrow\)}
\newcommand{\sbhead}{$S_{\mathrm{b}}\uparrow$}

\begin{minipage}[t]{0.48\linewidth}
    \centering
    \abltitle{(a) Various vision encoders}
    \begin{tabularx}{\linewidth}{@{}>{\raggedright\arraybackslash}X>{\centering\arraybackslash}p{3.5em}>{\centering\arraybackslash}p{3.5em}@{}}
    \toprule 
    \rowcolor{gray!10}
    Model & \kidhead & \sbhead \\ 
    \midrule
    CLIP (ours)    & 6.926 & 0.8029\\
    MAE~\cite{he2022mae}            & 8.058 & 0.7567\\
    SigLIP~\cite{zhai2023siglip}         & 6.676 & 0.8093\\
    % SigLIP2        & 6.810 \\
    DINOv3~\cite{simeoni2025dinov3}         & \textbf{6.663} & \textbf{0.8159}\\
    \bottomrule
    \end{tabularx}
\end{minipage}%
\hfill
\begin{minipage}[t]{0.48\linewidth}
    \centering 
    \abltitle{(b) Various MLLMs}
    \begin{tabularx}{\linewidth}{@{}>{\raggedright\arraybackslash}X>{\centering\arraybackslash}p{3.5em}>{\centering\arraybackslash}p{3.5em}@{}}
    \toprule
    \rowcolor{gray!10}
    Model & \kidhead & \sbhead \\ 
    \midrule
    LLaVA-1.6 (ours) & 6.926 & 0.8029\\
    BLIP-2~\cite{li2023blip}           & 7.132 & 0.8022\\
    InstructBLIP~\cite{dai2023instructblip}   & 6.841 & 0.8023\\ 
    Qwen2.5-VL~\cite{qwen25vl}       & \textbf{6.632} & \textbf{0.8031}\\
    \bottomrule
    \end{tabularx}
\end{minipage}

\endgroup
\end{minipage}

\end{table*}

\noindent \textbf{Effectiveness of the Intrinsic Appearance Module.}
Tab.~\ref{tab:abl_iam} examines how IAM extracts appearance cues from the foreground. High-pass filtering achieves the best overall result, suggesting that distributed high-frequency texture variations are more useful for appearance assimilation than sparse edge responses. The cumulative analysis shows that LAB color modeling provides the largest improvement in boundary concealment, while Gaussian smoothing stabilizes chromatic structure and the texture branch further refines local continuity. Together, these results confirm that IAM benefits from combining color guidance with texture cues to reduce appearance discrepancies between the target and the generated background.

% \begin{table}[!t]
% \centering
% \caption{Overall framework ablation of FreeCam. The table isolates CRM and the color and texture branches of IAM. KID is scaled by $10^{3}$. Lower KID and higher $S_{\mathrm{b}}$ are better. \textbf{Bold} indicates the best value in each reported metric.}
% \label{tab:abl_overall}
% \begingroup
% \footnotesize
% \setlength{\tabcolsep}{2pt}
% \renewcommand{\arraystretch}{1.10}
% \newcommand{\cmark}{\ding{51}}
% \newcommand{\xmark}{\ding{55}}
% \newcommand{\kidhead}{KID\(\downarrow\)}
% \newcommand{\sbhead}{$S_{\mathrm{b}}\uparrow$}
% \begin{tabular}{@{}>{\centering\arraybackslash}p{3.4em}>{\centering\arraybackslash}p{5.2em}>{\centering\arraybackslash}p{6.4em}>{\centering\arraybackslash}p{3.0em}>{\centering\arraybackslash}p{3.4em}@{}}
% \toprule
% \rowcolor{gray!10}
% CRM & IAM-Color & IAM-Texture & \kidhead & \sbhead \\
% \midrule
% \xmark & \xmark & \xmark & 13.326 & 0.2421\\
% \cmark & \xmark & \xmark & 8.244 & 0.6439\\
% \xmark & \cmark & \xmark & 8.981 & 0.7096\\
% \xmark & \xmark & \cmark & 11.424 & 0.6818\\
% \cmark & \cmark & \cmark & \textbf{6.926} & \textbf{0.8029}\\
% \bottomrule
% \end{tabular}
% \endgroup
% \end{table}

\noindent \textbf{Sensitivity of conditioning design.}
We further analyze the sensitivity of the conditioning design. For IAM, Fig.~\ref{fig:abl}(a) and Fig.~\ref{fig:abl}(b) quantitatively show that, across the evaluated design variants, intermediate CLIP features and the 10th denoising step provide the best balance between appearance assimilation and layout preservation. This is consistent with the role of IAM: shallow features or premature injection introduce noisy local constraints before the scene structure is stabilized, whereas overly semantic features or delayed injection weaken the color and texture cues needed for appearance assimilation between the foreground and its surroundings. The qualitative comparisons in Fig.~\ref{fig:clip_layer} and Fig.~\ref{fig:start_t} support the same conclusion, showing improved local continuity without disrupting the generated scene layout.

For CRM, Fig.~\ref{fig:abl}(c) shows that $K=5$ provides a balanced environment search space under our evaluated experimental settings. A smaller candidate set may miss camouflage-compatible contexts, while a larger one introduces semantically loose or visually incompatible environments. This reflects a trade-off between candidate diversity and relevance within the resulting environment search space. This supports the CRM design in which MLLM reasoning proposes a compact set of plausible environments and CLIP reranking selects the one most compatible with the foreground.

Fig.~\ref{fig:abl}(d) further studies where the appearance condition should be injected into the frozen U-Net. UpBlock-2 performs best because it lies at an intermediate decoding stage where the global background layout has emerged but local texture and color patterns remain adjustable throughout the remaining denoising steps. Injecting at higher-resolution blocks imposes appearance constraints locally, which can over-constrain background synthesis and reduce coherence.

% \noindent \textbf{Sensitivity of CRM search space.}
% Regarding CRM, we vary the number of candidate environments $K$ proposed by the MLLM, as shown in Fig.~\ref{fig:abl}(c). A small $K$ restricts the search space and may miss suitable camouflage contexts, whereas a large $K$ introduces semantically loose or visually incompatible candidates. The best result is achieved at $K=5$, which provides a balanced environment search space.

\begin{table*}[!t]
  \centering
  \caption{Camouflaged object detection results on four COD benchmarks under different synthetic-data augmentation ratios. Higher $S_{\alpha}$, $E_{\phi}$, and $F_{\beta}^{w}$ and lower $M$ are better. \textbf{Bold} and \underline{underlined} values denote the best and second-best results within each backbone, respectively.}
  \label{tab:sota_comparison_vertical}
  
  % 1. 极限压缩列间距
  \setlength{\tabcolsep}{1.5pt}
  
  % 2. 使用 resizebox 强制适应页面宽度
  \resizebox{\textwidth}{!}{%
    \begin{tabular}{l|cccc|cccc|cccc|cccc}
      \toprule
      % 第一行表头背景色
      \rowcolor{gray!10}
       & \multicolumn{4}{c|}{CHAMELEON} & \multicolumn{4}{c|}{CAMO} & \multicolumn{4}{c|}{COD10K} & \multicolumn{4}{c}{NC4K} \\
      
    %   \cmidrule(lr){2-5} \cmidrule(lr){6-9} \cmidrule(lr){10-13} \cmidrule(l){14-17}
      
      % 第二行表头背景色
      \rowcolor{gray!10}
       \multirow{-2}{*}{\textbf{Methods}} & $S_{\alpha}\uparrow$ & $E_{\phi}\uparrow$ & $F_{\beta}^{w}\uparrow$ & $M\downarrow$ 
       & $S_{\alpha}\uparrow$ & $E_{\phi}\uparrow$ & $F_{\beta}^{w}\uparrow$ & $M\downarrow$ 
       & $S_{\alpha}\uparrow$ & $E_{\phi}\uparrow$ & $F_{\beta}^{w}\uparrow$ & $M\downarrow$ 
       & $S_{\alpha}\uparrow$ & $E_{\phi}\uparrow$ & $F_{\beta}^{w}\uparrow$ & $M\downarrow$ \\
      \midrule
      \rowcolor{gray!10}
      \multicolumn{17}{c}{SINetv2~\cite{fan2021concealed}}
      \\

      Base
      & 0.881 & 0.933 & 0.796 & 0.035 
      & 0.806 & 0.858 & 0.708 & 0.081 
      & 0.802 & 0.876 & 0.656 & 0.040 
      & 0.837 & 0.887 & 0.746 & 0.053 \\
      % 修复点：添加 $...$ 包裹下标
      +LAKE-RED$_{100\%}$ 
      & 0.886 & 0.934 & 0.807 & 0.033 
      & 0.811 & 0.869 & 0.727 & 0.078 
      & 0.806 & 0.881 & 0.661 & 0.038 
      & 0.842 & 0.901 & 0.763 & 0.049 \\

      +FreeCam$_{10\%}$
      & 0.890 & 0.940 & 0.811 & 0.029 
      & 0.818 & 0.875 & 0.728 & 0.070 
      & 0.812 & 0.885 & 0.676 & 0.031 
      & 0.848 & 0.903 & 0.767 & 0.042 \\
      
      % 修复点：添加 $...$ 包裹下标
      +FreeCam$_{30\%}$ 
      & 0.894 & 0.952 & 0.822 & \underline{0.027} 
      & 0.829 & 0.884 & 0.739 & 0.067 
      & 0.818 & 0.890 & \underline{0.684} & 0.027 
      & \underline{0.856} & 0.909 & \underline{0.772} & \underline{0.036} \\
      
      % 修复点：添加 $...$ 包裹下标
      +FreeCam$_{50\%}$ 
      & \textbf{0.899} & \underline{0.957} & \underline{0.823} & \textbf{0.024} 
      & \textbf{0.834} & \underline{0.886} & \underline{0.744} & \underline{0.064} 
      & \textbf{0.820} & \underline{0.891} & 0.683 & \underline{0.026} 
      & \textbf{0.859} & \textbf{0.913} & \textbf{0.774} & \textbf{0.035} \\
      
      % 修复点：添加 $...$ 包裹下标
      +FreeCam$_{100\%}$ 
      & \underline{0.897} & \textbf{0.959} & \textbf{0.825} & \textbf{0.024} 
      & \underline{0.832} & \textbf{0.888} & \textbf{0.746} & \textbf{0.062} 
      & \underline{0.819} & \textbf{0.892} & \textbf{0.685} & \textbf{0.025} 
      & \underline{0.856} & \underline{0.910} & \textbf{0.774} & 0.038 \\
      \midrule
      \rowcolor{gray!10}
      \multicolumn{17}{c}{ESCNet~\cite{ESCNet}}
      \\
      
            Base 
      & 0.910 & 0.952 & 0.870 & 0.024 
      & 0.882 & 0.935 & 0.846 & 0.042 
      & 0.880 & 0.941 & 0.808 & 0.021 
      & 0.898 & 0.943 & 0.861 & 0.022 \\
      +LAKE-RED$_{100\%}$ 
      & 0.913 & 0.957 & 0.881 & 0.022 
      & 0.889 & 0.940 & 0.858 & 0.036 
      & 0.885 & 0.944 & 0.821 & \underline{0.019} 
      & 0.907 & 0.946 & 0.865 & 0.020 \\
      +FreeCam$_{10\%}$ 
      & 0.921 & 0.955 & 0.882 & 0.020 
      & 0.897 & 0.936 & 0.867 & \underline{0.035} 
      & 0.891 & 0.945 & 0.839 & \textbf{0.018} 
      & \underline{0.913} & 0.949 & 0.873 & 0.021 \\
      +FreeCam$_{30\%}$ 
      & \textbf{0.926} & \underline{0.958} & 0.884 & \underline{0.019} 
      & 0.903 & \underline{0.948} & 0.869 & 0.037 
      & \textbf{0.900} & \underline{0.948} & 0.833 & \underline{0.019} 
      & 0.911 & \underline{0.954} & \textbf{0.881} & \underline{0.019} \\
      +FreeCam$_{50\%}$ 
      & \underline{0.923} & 0.953 & \textbf{0.894} & 0.020 
      & \textbf{0.916} & 0.942 & \textbf{0.875} & \underline{0.035} 
      & \underline{0.898} & \textbf{0.949} & \textbf{0.842} & 0.021 
      & \textbf{0.914} & \textbf{0.955} & 0.876 & \underline{0.019} \\
      +FreeCam$_{100\%}$ 
      & 0.919 & \textbf{0.959} & \underline{0.891} & \textbf{0.018} 
      & \underline{0.907} & \textbf{0.950} & \underline{0.871} & \textbf{0.032} 
      & \underline{0.898} & \textbf{0.949} & \underline{0.841} & \textbf{0.018} 
      & 0.908 & \underline{0.954} & \underline{0.877} & \textbf{0.017} \\
      \bottomrule
    \end{tabular}%
  }
\end{table*}

\begin{table}[!t]
  \centering
  \caption{Detection degradation on the overall set and three subsets using FreeCam$_{\mathrm{SD1.5}}$. The \textit{degradation} rows report changes from Original to FreeCam for each evaluated detector. \textcolor{brickred}{Red decreases} and \textcolor{oceanblue}{blue increases} indicate better and worse camouflage, respectively.}
  \label{tab:sd15_detection_degradation}

  \begingroup
%   \small
  \renewcommand{\arraystretch}{1.22}
  \setlength{\tabcolsep}{1.8pt}
  \newcommand{\detdrop}[1]{\textcolor{brickred}{$\downarrow #1$}}
  \newcommand{\detrise}[1]{\textcolor{oceanblue}{$\uparrow #1$}}
  \resizebox{\linewidth}{!}{%
    \begin{tabular}{l|ccc|cc|c}
      \toprule
      \rowcolor{gray!10}
      & \multicolumn{3}{c|}{\textbf{Average Precision}} & \multicolumn{2}{c|}{\textbf{Recall}} & \textbf{Overlap} \\
      \rowcolor{gray!10}
      \multirow{-2}{*}{\textbf{Method}}
      & $\mathrm{AP}_{50}\uparrow$ & $\mathrm{AP}_{75}\uparrow$ & $\mathrm{mAP}\uparrow$ & $\mathrm{R}_{50}\uparrow$ & $\mathrm{R}_{75}\uparrow$ & $\mathrm{IoU}\uparrow$ \\
      \midrule
      \rowcolor{gray!12}
      \multicolumn{7}{l}{\textbf{Overall}} \\
      YOLOv10-X & 0.393 & 0.348 & 0.332 & 0.680 & 0.614 & 0.638 \\
      +FreeCam$_{\mathrm{SD1.5}}$ & 0.372 & 0.224 & 0.224 & 0.533 & 0.383 & 0.473 \\
      \rowcolor{best}
      \textit{degradation} & \detdrop{0.021} & \detdrop{0.124} & \detdrop{0.108} & \detdrop{0.147} & \detdrop{0.231} & \detdrop{0.165} \\
      \cmidrule(lr){1-7}
      G-DINO-B & 0.394 & 0.331 & 0.311 & 0.785 & 0.712 & 0.738 \\
      +FreeCam$_{\mathrm{SD1.5}}$ & 0.361 & 0.169 & 0.186 & 0.623 & 0.420 & 0.562 \\
      \rowcolor{best}
      \textit{degradation} & \detdrop{0.033} & \detdrop{0.162} & \detdrop{0.125} & \detdrop{0.162} & \detdrop{0.292} & \detdrop{0.176} \\
      \midrule
      \rowcolor{gray!12}
      \multicolumn{7}{l}{\textbf{CO}} \\
      YOLOv10-X & 0.253 & 0.169 & 0.166 & 0.359 & 0.278 & 0.335 \\
      +FreeCam$_{\mathrm{SD1.5}}$ & 0.121 & 0.033 & 0.048 & 0.217 & 0.106 & 0.200 \\
      \rowcolor{best}
      \textit{degradation} & \detdrop{0.132} & \detdrop{0.136} & \detdrop{0.118} & \detdrop{0.142} & \detdrop{0.172} & \detdrop{0.135} \\
      \cmidrule(lr){1-7}
      G-DINO-B & 0.665 & 0.564 & 0.528 & 0.861 & 0.778 & 0.809 \\
      +FreeCam$_{\mathrm{SD1.5}}$ & 0.266 & 0.076 & 0.109 & 0.498 & 0.257 & 0.468 \\
      \rowcolor{best}
      \textit{degradation} & \detdrop{0.399} & \detdrop{0.488} & \detdrop{0.419} & \detdrop{0.363} & \detdrop{0.521} & \detdrop{0.341} \\
      \midrule
      \rowcolor{gray!12}
      \multicolumn{7}{l}{\textbf{SO}} \\
      YOLOv10-X & 0.487 & 0.392 & 0.380 & 0.700 & 0.607 & 0.652 \\
      +FreeCam$_{\mathrm{SD1.5}}$ & 0.433 & 0.237 & 0.247 & 0.578 & 0.400 & 0.509 \\
      \rowcolor{best}
      \textit{degradation} & \detdrop{0.054} & \detdrop{0.155} & \detdrop{0.133} & \detdrop{0.122} & \detdrop{0.207} & \detdrop{0.143} \\
      \cmidrule(lr){1-7}
      G-DINO-B & 0.376 & 0.274 & 0.269 & 0.713 & 0.606 & 0.659 \\
      +FreeCam$_{\mathrm{SD1.5}}$ & 0.381 & 0.172 & 0.192 & 0.656 & 0.442 & 0.576 \\
      \rowcolor{best}
      \textit{degradation} & \detrise{0.005} & \detdrop{0.102} & \detdrop{0.077} & \detdrop{0.057} & \detdrop{0.164} & \detdrop{0.083} \\
      \midrule
      \rowcolor{gray!12}
      \multicolumn{7}{l}{\textbf{GO}} \\
      YOLOv10-X & 0.583 & 0.565 & 0.535 & 0.982 & 0.958 & 0.927 \\
      +FreeCam$_{\mathrm{SD1.5}}$ & 0.571 & 0.403 & 0.380 & 0.805 & 0.643 & 0.710 \\
      \rowcolor{best}
      \textit{degradation} & \detdrop{0.012} & \detdrop{0.162} & \detdrop{0.155} & \detdrop{0.177} & \detdrop{0.315} & \detdrop{0.217} \\
      \cmidrule(lr){1-7}
      G-DINO-B & 0.493 & 0.274 & 0.250 & 0.782 & 0.751 & 0.747 \\
      +FreeCam$_{\mathrm{SD1.5}}$ & 0.437 & 0.271 & 0.266 & 0.715 & 0.561 & 0.640 \\
      \rowcolor{best}
      \textit{degradation} & \detdrop{0.056} & \detdrop{0.003} & \detrise{0.016} & \detdrop{0.067} & \detdrop{0.190} & \detdrop{0.107} \\
      \bottomrule
    \end{tabular}%
  }
  \endgroup
\end{table}

\begin{figure}[!t]
\centering
\includegraphics[width=0.85\linewidth]{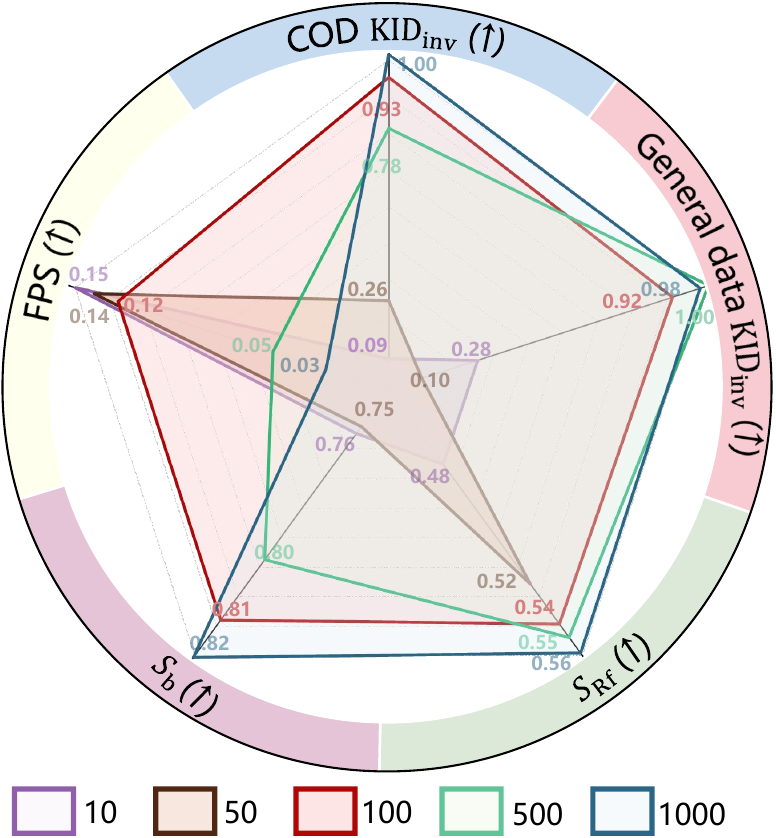}
% \caption{Environment-set scale analysis in CRM. Candidate pools of 10, 50, 100, 500, and 1000 environments are compared across five normalized axes. The KID-related axes are inverse-normalized so that larger values indicate better performance.}
\caption{Effect of the CRM environment-set size across five normalized metrics. KID axes are inverse-normalized so that higher values are better.}
\label{fig:envset}
\end{figure}

\noindent \textbf{Model-choice sensitivity.}
Tab.~\ref{tab:abl_model_choice} examines whether FreeCam depends on a specific foundation component. The default CLIP and LLaVA-1.6 setting is used for the main experiments to provide a reproducible and widely comparable configuration, rather than because it is numerically optimal. Replacing CLIP with stronger visual encoders such as SigLIP~\cite{zhai2023siglip} or DINOv3~\cite{simeoni2025dinov3} improves the appearance branch, suggesting that IAM benefits from richer foreground representations for color and texture guidance. Similarly, replacing LLaVA-1.6 with Qwen2.5-VL~\cite{qwen25vl} improves the reasoning branch, indicating that CRM can leverage stronger multimodal reasoning to propose more suitable camouflage contexts. These results support the modularity of FreeCam: its semantic-appearance conditioning is not tied to a particular encoder or MLLM, and can naturally benefit from advances in off-the-shelf foundation models without CIG-specific training.

\begin{figure*}[!t]
\centering
\includegraphics[width=0.92\linewidth]{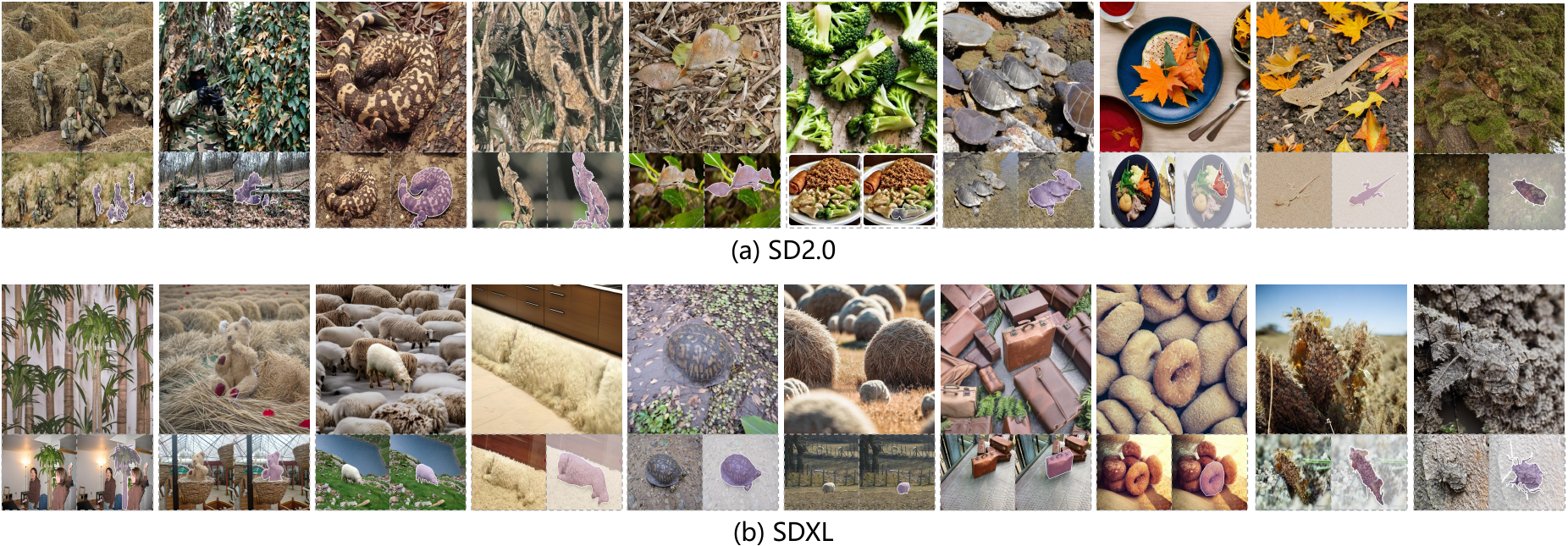}
\caption{Generation results across different frozen diffusion backbones. FreeCam is applied to (a) SD2.0 and (b) SDXL using the same training-free formulation, illustrating that its semantic and appearance conditioning can be transferred across pretrained generative models.}
\label{fig:arch}
\end{figure*}

\noindent \textbf{Environment-set scale.}
Fig.~\ref{fig:envset} analyzes the effect of the CRM environment-set size on generation quality, camouflage effectiveness, and efficiency. Across all evaluated scales, the candidate environments are manually curated from Places365 scene categories~\cite{zhou2017places} and publicly available web sources. For visualization, the KID axes are inverse-normalized so that larger values indicate better performance, consistent with $S_{\mathrm{Rf}}$, $S_{\mathrm{b}}$, and FPS. A small environment set is efficient, but its limited semantic coverage may miss suitable hiding contexts, leading to weaker KID and appearance assimilation. Increasing the set size generally improves generation quality and camouflage effectiveness, as CRM can search from a richer pool of environments compatible with the target and favorable to concealment. However, the improvement comes with a clear efficiency cost, and overly large sets may introduce weakly related categories that dilute the concealment-oriented search space. The 100-category setting therefore provides a practical balance, preserving strong generation quality and camouflage effectiveness while avoiding the efficiency loss caused by large candidate pools.

% \FloatBarrier

\subsection{Downstream Object Detection}

Beyond the above analysis, we further assess FreeCam through complementary downstream settings from a task-level perspective. First, synthesized samples augment camouflaged object detection to test whether they provide useful camouflage variations for training. Second, we compare general object detectors before and after FreeCam generation to determine whether the synthesized surroundings reduce target detectability. Together, these settings assess whether FreeCam improves camouflage-aware learning and makes foreground objects harder to localize, thereby producing functional camouflage rather than merely plausible inpainted backgrounds.

\noindent \textbf{COD augmentation protocol.}
For COD augmentation, we train the real-data baseline on 4,040 images from the COD10K training set~\cite{fan2020camouflaged}. We then augment this set with FreeCam-generated samples at ratios of 10\%, 30\%, 50\%, and 100\%, producing training sets ranging from 4,444 to 8,080 images. This protocol tests whether concealment-oriented synthetic samples can complement real COD data under controlled augmentation ratios. For SINetv2~\cite{fan2021concealed}, we use Adam~\cite{kingma2014adam} with an initial learning rate of \(1\times10^{-4}\), a decay factor of 0.1 every 50 epochs, 100 training epochs, batch size 16, and an input resolution of \(352\times352\). Gradient clipping with a maximum norm of 0.5 is applied for stable optimization, and the loss combines weighted binary cross-entropy with weighted IoU over multi-scale predictions. ESCNet~\cite{ESCNet} follows the same augmentation splits for consistency.

\noindent \textbf{Camouflaged object detection augmentation.} 
Tab.~\ref{tab:sota_comparison_vertical} evaluates whether FreeCam can serve as a data engine for COD training. Across SINetv2~\cite{fan2021concealed} and ESCNet~\cite{ESCNet}, adding FreeCam-generated samples improves the real-data baseline on CHAMELEON~\cite{skurowski2018animal}, CAMO~\cite{le2019anabranch}, COD10K~\cite{fan2020camouflaged}, and NC4K~\cite{lv2021simultaneously} in most metrics, with stronger gains typically observed at moderate or large augmentation ratios. The improvement over both the original training set and LAKE-RED augmentation suggests that the benefit does not simply come from increasing the number of training images. Instead, FreeCam introduces concealment-oriented variations: CRM expands the range of semantically compatible hiding contexts, while IAM encourages foreground-background appearance assimilation through color and texture cues. These samples therefore complement the limited scene and appearance coverage of real COD data. The gains are not strictly monotonic with the augmentation ratio, since synthetic data also changes the training distribution. Nevertheless, the consistent improvements across two detectors and four benchmarks indicate that FreeCam-generated data provide useful supervision for camouflage-aware recognition.

\noindent \textbf{General object detection degradation.}
Tab.~\ref{tab:sd15_detection_degradation} evaluates FreeCam from the perspective of independent object detectors. We compare YOLOv10-X~\cite{wang2024yolov10} and G-DINO-B~\cite{liu2024grounding} on the original images and their FreeCam$_{\mathrm{SD1.5}}$ counterparts. After FreeCam generation, almost all AP, recall, and IoU scores decrease, with particularly large drops on CO and consistent recall and overlap degradation across SO and GO. The simultaneous decrease in recall and overlap is important, as it indicates that the generated camouflage reduces both target discoverability and localization quality. This suggests that FreeCam affects not only distribution-level image quality, but also the perceptual cues used by recognition systems to localize objects. This trend is consistent with FreeCam's design: CRM provides a plausible hiding context and IAM weakens foreground-background appearance separability, although the few metric increases on non-COD subsets indicate that detector degradation should be treated as complementary rather than standalone evidence of camouflage quality.

\subsection{Further Analysis}

\noindent \textbf{Generalization across base models.}
To assess cross-backbone scalability, we instantiate FreeCam on stronger frozen diffusion priors without CIG-specific parameter updates. As shown in Fig.~\ref{fig:arch}, the same framework preserves plausible concealment when the generative backbone is changed. This behavior follows from the formulation of FreeCam: camouflage guidance is expressed as semantic and appearance conditions and injected through frozen cross-attention, rather than learned as backbone-specific denoising weights. As a result, stronger generative priors can improve visual fidelity while the concealment objective remains governed by the same guidance mechanism.

\begin{figure}[!t]
\centering
\includegraphics[width=0.98\linewidth]{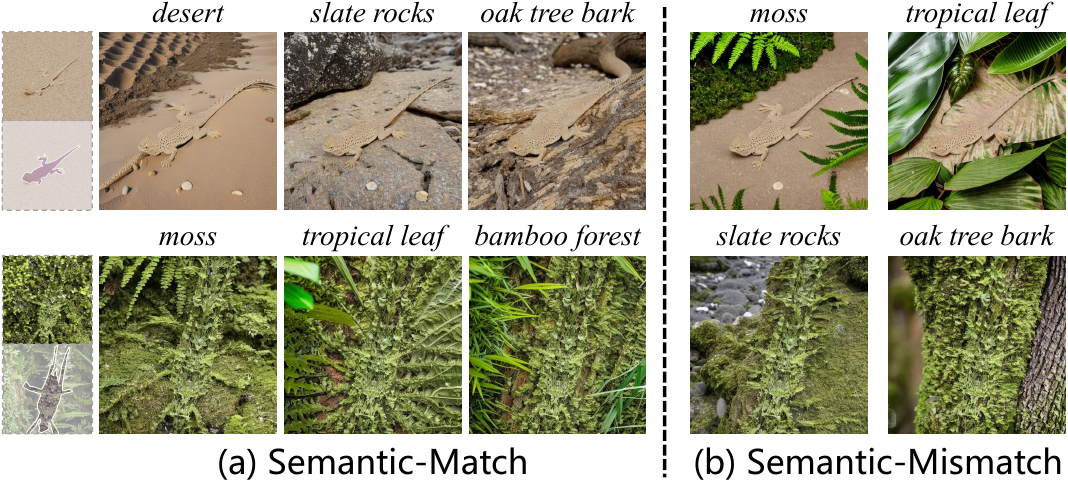}
\caption{Controllability analysis under user-designated environments. (a) Semantic-match settings use environments compatible with the foreground object, whereas (b) semantic-mismatch settings force less compatible contexts to test controllability.} 
\label{fig:control}
\end{figure}

\begin{figure}[!t]
\centering
\includegraphics[width=0.98\linewidth]{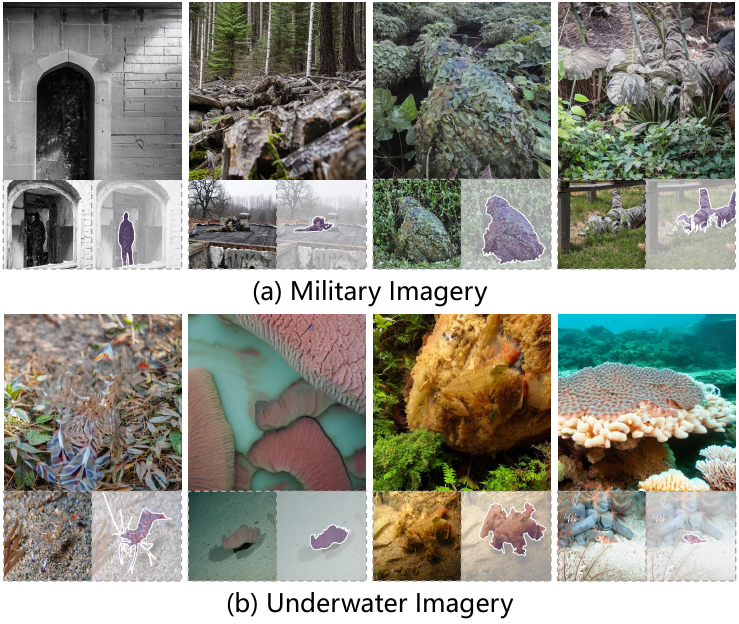}
\caption{Cross-scenario generalization of FreeCam. The examples cover (a) military imagery and (b) underwater imagery, showing camouflage synthesis beyond the COD-style evaluation setting.} 
\label{fig:cross}
\end{figure}

\noindent \textbf{Controllability analysis.}
To evaluate the controllability of our method, we restrict the environment set size to $1$ and the retrieval parameter $K=1$, forcing the model to synthesize camouflage strictly conditioned on a specific semantic context, as shown in Fig.~\ref{fig:control}. Here, \emph{Semantic-Match} indicates an environment compatible with the foreground, while \emph{Semantic-Mismatch} means an incompatible one. The results indicate that the proposed FreeCam can significantly expand generative versatility, enabling synthesis of camouflage patterns tailored to user-designated environments from the candidate set.

\noindent \textbf{Generalization to diverse scenarios.}
Fig.~\ref{fig:cross} qualitatively examines FreeCam under scenarios beyond the standard COD evaluation setting, including military and underwater environments with distinct visual characteristics. The results show that FreeCam can maintain coherent camouflage in complex visual environments because the generated context is selected according to foreground-aware semantic compatibility and further refined by intrinsic appearance guidance. This suggests that FreeCam transfers the frozen inpainting prior to concealment-oriented synthesis by deriving the semantic context and appearance condition from the foreground itself.

\begin{figure}[!t]
\centering
\includegraphics[width=0.98\linewidth]{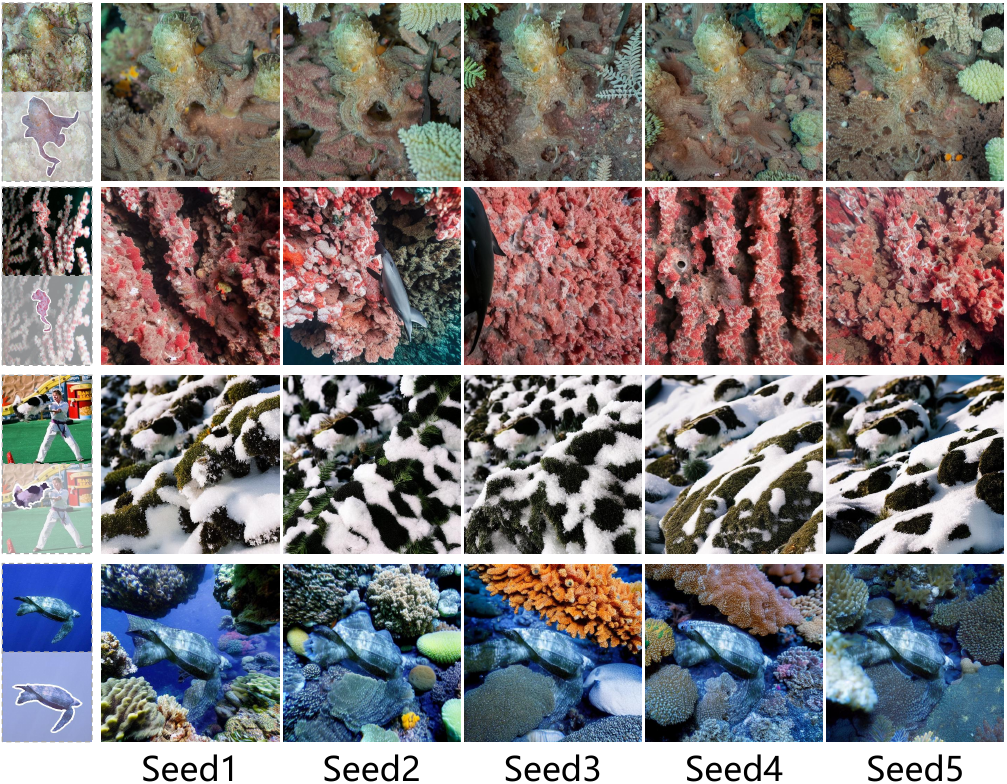}
\caption{Random-seed sensitivity of FreeCam. For each input, five outputs are generated under fixed semantic and appearance conditions using different random seeds.}
\label{fig:seed}
\end{figure}

\noindent \textbf{Random-seed stability.}
We further examine the sampling stability of FreeCam under different random seeds, as shown in Fig.~\ref{fig:seed}. Although the frozen diffusion sampler introduces stochastic variation in the synthesized background, the generated results remain guided by the same foreground-derived semantic and appearance conditions. The visual differences are therefore mainly reflected in local layout and texture details, while the intended concealment conditions and appearance assimilation are effectively preserved.

\begin{figure}[!t]
\centering
\includegraphics[width=0.98\linewidth]{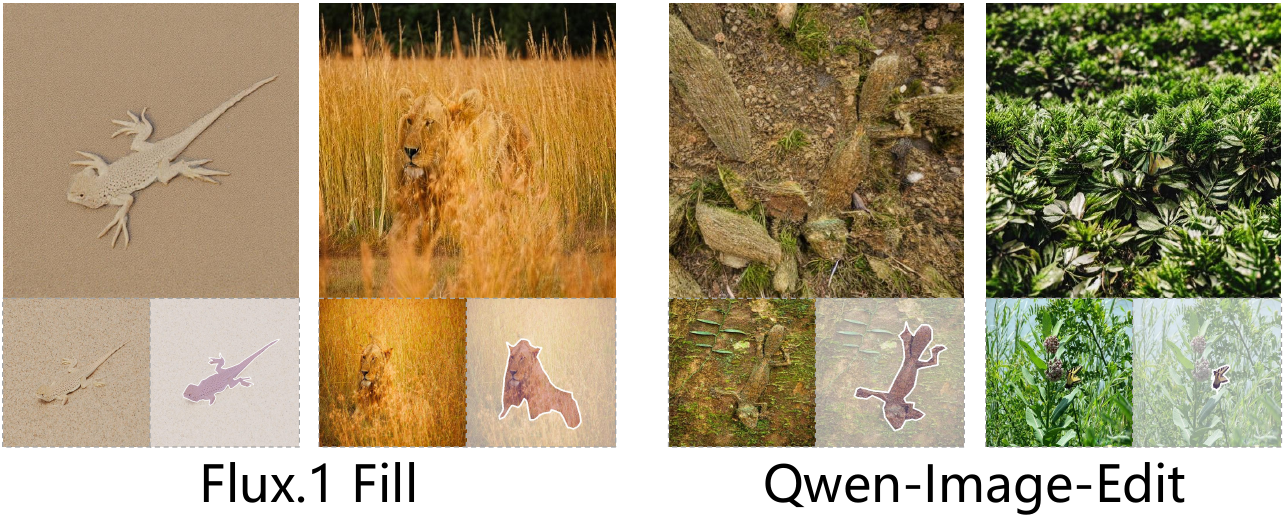}
\caption{Feasibility of FreeCam on larger image-editing models. The comparison shows the examples generated with FLUX.1 Fill~\cite{flux2024} and Qwen-Image-Edit~\cite{qwenimage}.}
\label{fig:LM}
\end{figure}

\begin{figure}[!t]
\centering
\includegraphics[width=0.98\linewidth]{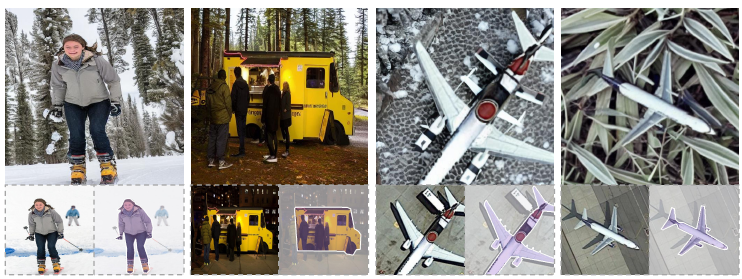}
\caption{Limitation analysis. The examples show typical failure cases, mainly involving highly salient targets such as human targets and aircraft in remote sensing imagery.} 
\label{fig:limit}
\end{figure}

\noindent \textbf{Large-model feasibility.}
Fig.~\ref{fig:LM} explores whether the FreeCam protocol can be transferred to larger image-editing foundations. The purpose is not to claim a new quantitative advantage, but to examine whether concealment-oriented guidance can remain valid when the underlying editing prior is replaced. The results show that foreground-derived semantic context and appearance cues can still guide large editing models toward camouflage synthesis at inference time without CIG-specific training. This supports the formulation of FreeCam as a training-free CIG paradigm rather than a specific implementation.

\noindent \textbf{Limitations.}
Despite its flexibility, FreeCam still exhibits failure modes, as shown in Fig.~\ref{fig:limit}. The targets with distinctive semantic or geometric structures may remain salient even when the generated background is semantically compatible and visually consistent with the foreground. These cases expose an inherent tension between foreground preservation and concealment: FreeCam retains the original target, while CRM and IAM regulate only the surrounding context and its appearance relations. Consequently, separability caused primarily by distinctive object geometry cannot be reduced through background synthesis alone.

% \FloatBarrier

\section{Conclusion}
% In this work, we formulate training-free CIG and instantiate it with FreeCam. FreeCam preserves the original foreground by restricting synthesis to the background region, while CRM and IAM promote semantic compatibility and appearance assimilation, respectively. Experiments demonstrate strong generation quality, improved camouflage effectiveness, and lower deployment overhead than training-based CIG methods. Downstream results further show the utility of the generated images for camouflaged object detection. These findings support training-free CIG as a practical alternative to COD-specific optimization, while the remaining failure cases highlight the tension between foreground preservation and concealment for structurally distinctive targets.

In this work, we formulate training-free CIG as a concealment-oriented paradigm and identify semantic compatibility and appearance assimilation as its two camouflage-specific challenges. FreeCam instantiates these principles through CRM and IAM, while background-only synthesis preserves the original foreground. Extensive experiments demonstrate strong generation quality, improved camouflage effectiveness, and lower deployment overhead than training-based CIG methods, with further benefits for downstream camouflaged object detection. These findings support training-free CIG as a practical direction for camouflage synthesis while highlighting the remaining tension between foreground preservation and concealment.

% In this work, we formulate camouflage image generation from a training-free perspective and introduce FreeCam as a concrete instantiation of this paradigm. By deriving semantic context and intrinsic appearance cues directly from the foreground, FreeCam guides a frozen inpainting diffusion model to synthesize backgrounds that are semantically compatible with the target and visually assimilated to its color and texture, while preserving the foreground by construction. Extensive experiments show that FreeCam achieves strong generation quality, improved appearance assimilation and boundary concealment, and lower deployment overhead than training-based CIG methods, with further benefits for downstream camouflaged object detection. These results suggest that effective CIG does not have to rely exclusively on COD-specific optimization, and support training-free CIG as a practical direction for camouflage synthesis.

\noindent \textbf{Data Availability.} The LAKE-RED dataset used in this study is publicly available through its official repository: \url{https://github.com/PanchengZhao/LAKE-RED}.

\bibliography{sn-bibliography}

\end{document}